\documentclass[letterpaper]{article} % DO NOT CHANGE THIS
\usepackage{aaai24}  % DO NOT CHANGE THIS
\usepackage{times}  % DO NOT CHANGE THIS
\usepackage{helvet}  % DO NOT CHANGE THIS
\usepackage{courier}  % DO NOT CHANGE THIS
\usepackage[hyphens]{url}  % DO NOT CHANGE THIS
\usepackage{graphicx} % DO NOT CHANGE THIS
\usepackage{natbib}  % DO NOT CHANGE THIS AND DO NOT ADD ANY OPTIONS TO IT
\usepackage{caption} % DO NOT CHANGE THIS AND DO NOT ADD ANY OPTIONS TO IT
\usepackage[
linesnumbered,ruled,vlined]{algorithm2e}
\usepackage {algpseudocode}
\usepackage{algorithmicx}
\usepackage{algcompatible}

\makeatletter
\newcommand{\setlabel}[1]{\edef\@currentlabel{#1}\label}
\makeatother

\usepackage{subfigure}
\usepackage{multirow}
\usepackage{amsmath, amssymb}
\usepackage{booktabs}
\usepackage{longtable}
\usepackage{makecell}
\usepackage{subcaption}
\usepackage[table]{xcolor}
\usepackage{tikz}
\newcommand{\tikzxmark}{%
\tikz[scale=0.23] {
    \draw[line width=0.7,line cap=round] (0,0) to [bend left=6] (1,1);
    \draw[line width=0.7,line cap=round] (0.2,0.95) to [bend right=3] (0.8,0.05);
}}
\newcommand{\tikzcmark}{%
\tikz[scale=0.23] {
    \draw[line width=0.7,line cap=round] (0.25,0) to [bend left=10] (1,1);
    \draw[line width=0.8,line cap=round] (0,0.35) to [bend right=1] (0.23,0);
}}

\title{Context Enhanced Transformer for Single Image Object Detection}
\author{
    Seungjun An\equalcontrib\textsuperscript{\rm 1},
    Seonghoon Park\equalcontrib\textsuperscript{\rm 1},
    Gyeongnyeon Kim\equalcontrib\textsuperscript{\rm 1},\\
    Jeongyeol Baek\textsuperscript{\rm 2},
    Byeongwon Lee\textsuperscript{\rm 2},
    Seungryong Kim\textsuperscript{\rm 1}
}
\affiliations{
    \textsuperscript{\rm 1}Korea University, Seoul, Korea\\
    \textsuperscript{\rm 2}SK Telecom, Seoul, Korea\\
    \{dkstmdwns, seong0905, kkn9975\}@korea.ac.kr, \{jeongyeol.baek, bwon.lee\}@sk.com, seungryong\_kim@korea.ac.kr
}

\usepackage{bibentry}
\begin{document}

\maketitle

\begin{abstract}
With the increasing importance of video data in real-world applications, there is a rising need for efficient object detection methods that utilize temporal information. While existing video object detection (VOD) techniques employ various strategies to address this challenge, they typically depend on locally adjacent frames or randomly sampled images within a clip. Although recent Transformer-based VOD methods have shown promising results, their reliance on multiple inputs and additional network complexity to incorporate temporal information limits their practical applicability. In this paper, we propose a novel approach to single image object detection, called Context Enhanced TRansformer (CETR), by incorporating temporal context into DETR using a newly designed memory module. To efficiently store temporal information, we construct a class-wise memory that collects contextual information across data. Additionally, we present a classification-based sampling technique to selectively utilize the relevant memory for the current image. In the testing, We introduce a test-time memory adaptation method that updates individual memory functions by considering the test distribution. Experiments with CityCam and ImageNet VID datasets exhibit the efficiency of the framework on various video systems. The project page and code will be made available at: \textcolor{magenta}{https://ku-cvlab.github.io/CETR}.
\end{abstract}

\section{Introduction}
Object detection is one of the fundamental and essential tasks in the computer vision field with its extensive versatility across a wide range of applications. Moreover, various applications in real-world scenarios, including video surveillance~\cite{nascimento2006performance, fu2019foreground}, autonomous driving~\cite{chen2015deepdriving,chen2016monocular}, and robot navigation~\cite{hernandez2016object}, heavily rely on video data. Despite the remarkable success of object detectors for a single image, directly applying them to video data encounters challenges due to appearance deterioration caused by motions and occlusions.
\begin{figure}[t]
\centering
\includegraphics[width=1.0\linewidth]{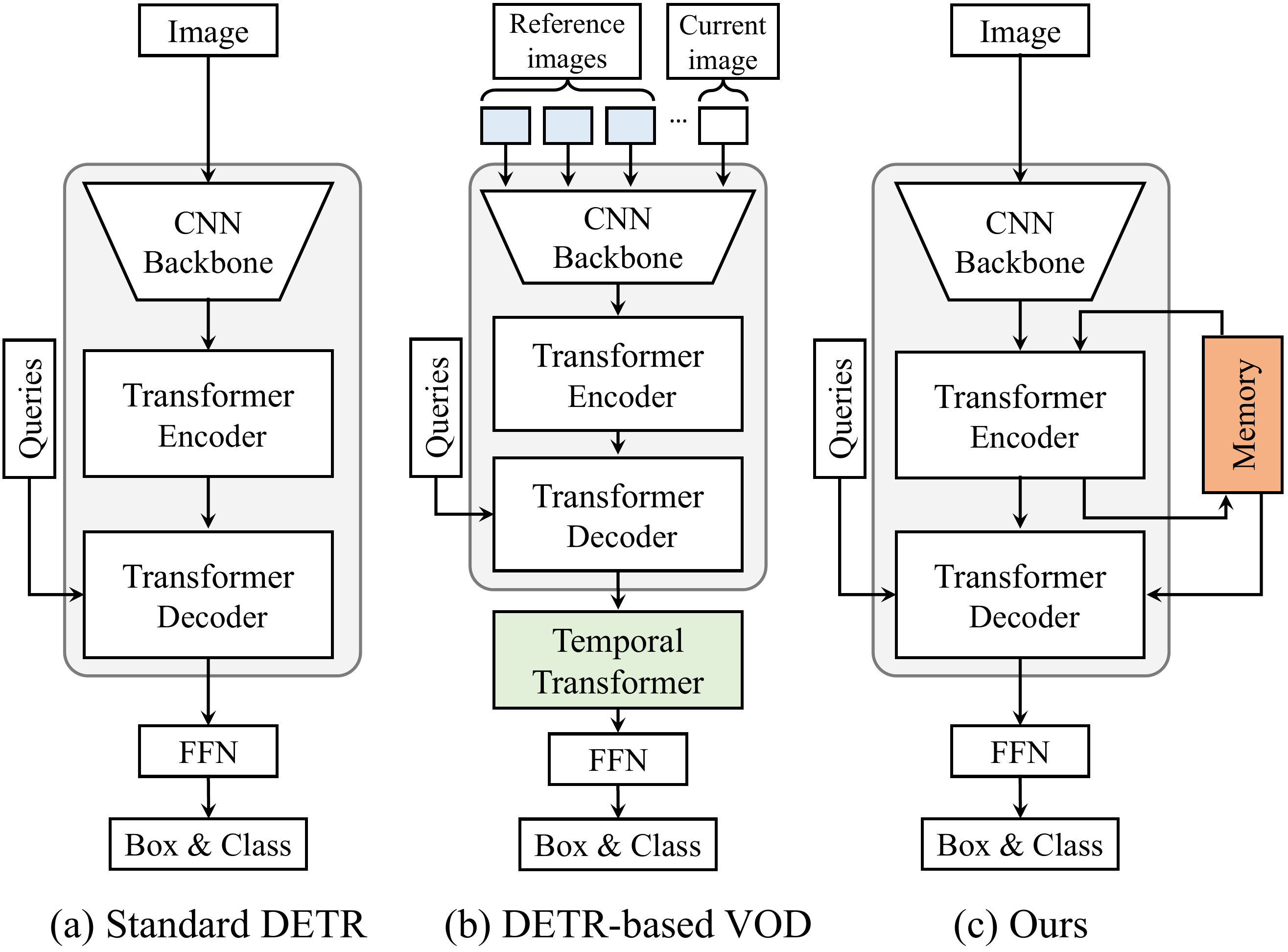}
\vspace{-15pt}
\caption{Comparisons between existing works and ours. (a) standard DETR~\cite{carion2020end}, (b) DETR-based video object detection~\cite{zhou2022transvod}, and (c) our proposed framework, dubbed CETR. Our method effectively detects objects in video data without adding heavy components. }
\label{fig:overview}
\vspace{-15pt}
\end{figure}

To address this challenge, video object detection (VOD) models~\cite{zhu2017flow, wang2018fully} have been proposed to improve object detection performance by leveraging temporal information. Previous approaches usually aggregate features from nearby frames exploiting optical flow~\cite{zhu2017flow, wang2018fully} or LSTM~\cite{kang2017object, kang2017t}. Nevertheless, these methods primarily focus on short-term frames, thus limiting their ability to capture a more extensive feature representation. 
To overcome this limitation, attention-based approaches~\cite{chen2020memory,deng2019object,deng2019relation} attempt to capture long-range temporal dependency by utilizing memory structures to aggregate features globally or locally. Yet, they depend on randomly sampled images within a clip, which struggle to integrate holistic contextual information from video data. Furthermore, the construction of stacked memory modules to store features of adjacent frames incurs high computational costs and unnecessary memory usage.

On the other hand, in light of the remarkable performance of Transformer-based models in image object detection~\cite{carion2020end, liu2022dab}, e.g., detection with Transformers (DETR), researchers have commenced extending them to the video domain~\cite{zhou2022transvod,wang2022ptseformer}.
However, these methods exhibit an essential reliance on auxiliary networks and the need for multiple sequential frames, as shown in Fig.~\ref{fig:overview}. Such prerequisites lead to a considerable decrease in processing speed, thereby failing to fulfill the real-time operational demands of various systems. As a consequence, there is a need for more efficient and streamlined approaches that can meet the real-time requirements essential for practical applications.

In this work, we propose a novel single image object detection method, dubbed Context Enhanced TRansformer (CETR), that effectively incorporates contextual information across the given data. Following the recent trend of Transformer-based detectors, we adopt DETR as our baseline. Due to the inherent attention mechanism within the Transformer framework, our approach effectively incorporates temporal information through attention modules.
In order to utilize temporal context without requiring additional reference frames or networks, we present a context memory module (CMM) that stores class-wise feature representations and updates effectively using momentum update in a non-parametric manner. The memory also represents each class as a set of prototypes, allowing intra-classes to contain a variety of attributes. 
In addition, to effectively capture relevant information for the current features, we introduce a score-based sampling methodology. By propagating the encoded memory features through a classification network for making predictions, CETR employs a sampled class-specific memory that closely aligns with the current input.  
Furthermore, we introduce an adaptive memory updating technique tailored to the test domain across different camera settings. Unlike the uniform exponential moving average update employed during training, we implement an online updating strategy aligned with the class-wise distribution of the test domain. 
Utilizing a weighted sum of the target and source domain memories, this strategy facilitates adaptation toward the test data distribution while retaining contextual information from the training phase.

To validate the effectiveness of the proposed method, we conduct extensive experiments on the CityCam dataset~\cite{zhang2017understanding}, one of the real traffic video data. Furthermore, experiments on the ImageNet VID~\cite{russakovsky2015imagenet} demonstrate that our framework achieves comparable accuracy with the state-of-the-art video object detectors with a much faster speed and efficient memory resource. We also perform detailed ablation studies and deeply analyze the memory module to confirm that it is effective at capturing contextual information.  
\begin{figure*}[t]
\begin{center}
\includegraphics[width=1\linewidth]{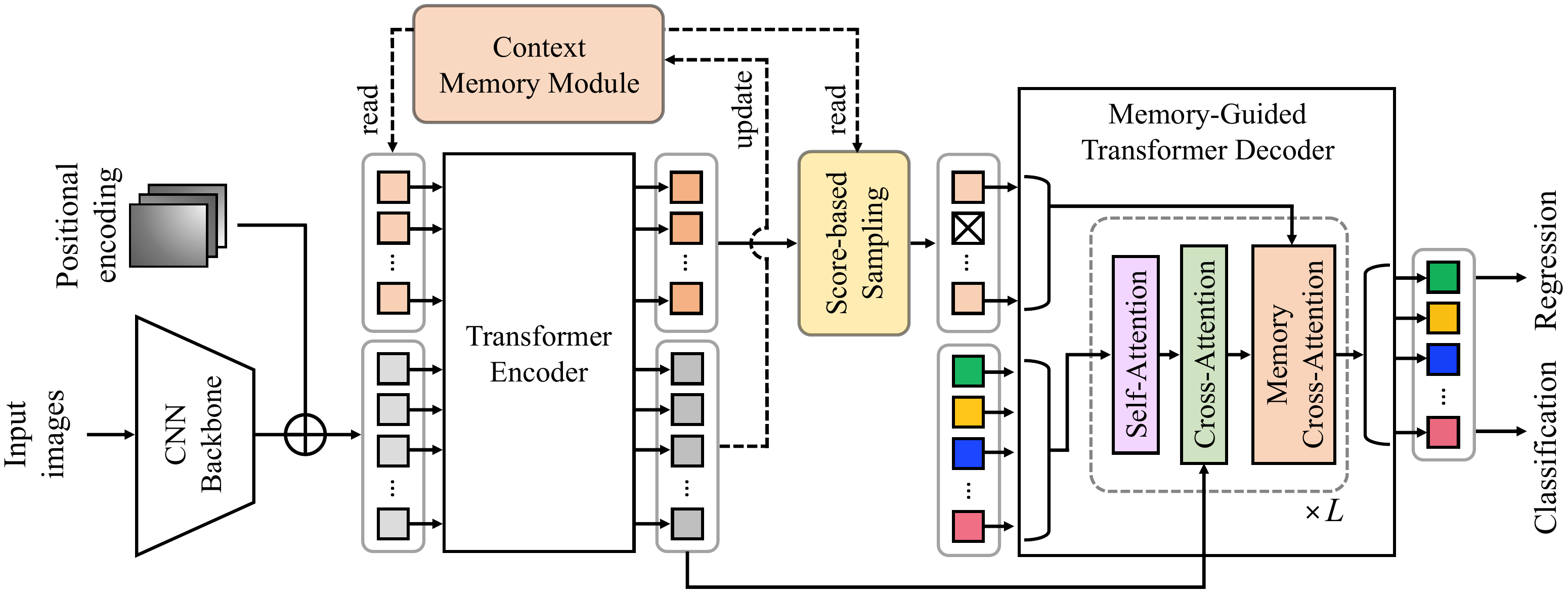}
\end{center}
\vspace{-10pt}
\caption{Overview of our framework. CETR builds upon the DETR~\cite{carion2020end} architecture. 
Within our framework, a pivotal component is the context memory module (CMM), which serves as the input for the Transformer encoder.
Subsequently, the encoded memory features are passed through the classification network. Predicted probability serves as a threshold for score-based sampling. The sampled class-wise memory is aggregated with the query using the cross-attention mechanism within the memory-guided Transformer decoder (MGD). }
\label{fig:main_figure}
\vspace{-15pt}
\end{figure*}
\begin{figure}[t]
\centering
\includegraphics[width=0.95\linewidth]{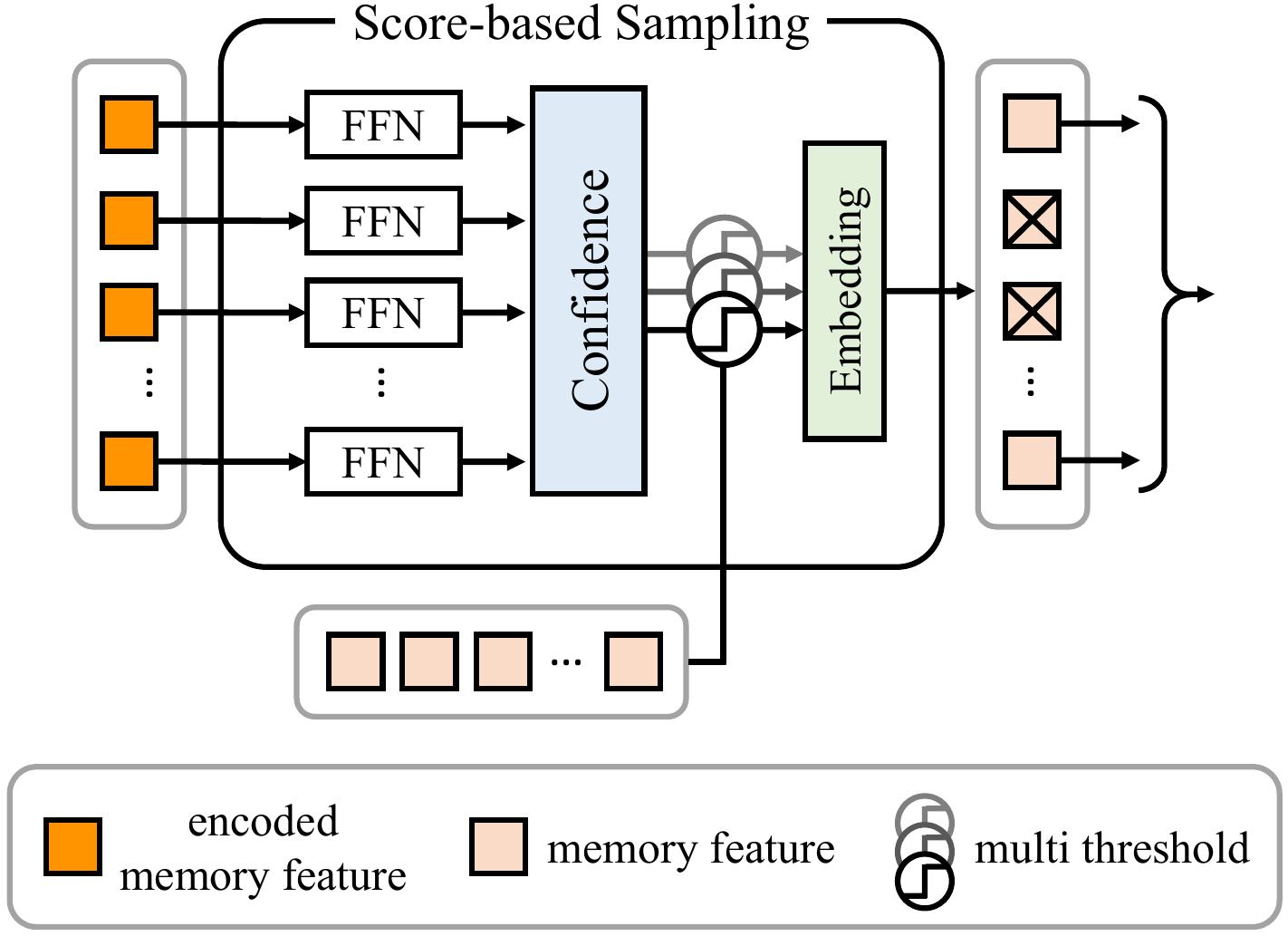}
\vspace{-5pt}
\caption{
     Details of score-based sampling module. } 
\label{fig:sampling}
\vspace{-15pt}
\end{figure}

\section{Related Work}
\paragraph{Single image object detection}
Single image object detectors have been extensively explored due to the development of deep convolutional
neural networks (CNNs). CNN-based object detectors can be classified into two pipelines: two-stage and one-stage detectors. Two-stage detectors~\cite{girshick2015fast, ren2015faster, dai2016r} generate coarse object proposals and then classify the proposals and regress the bounding boxes to refine them. In contrast, one-stage detectors~\cite{duan2019centernet,tian2019fcos} directly predict object locations and categories in an image by utilizing densely designed anchors. In recent years, DETR~\cite{carion2020end}, a prominent Transformer-based object detector, casts object detection as a direct set prediction problem by removing hand-crafted representations and post-processing techniques. Many follow-up works~\cite{li2022dn,liu2022dab,meng2021conditional,zhu2020deformable} have attempted to address the slow training convergence of DETR’s inefficient design and use of queries.  In this paper, we choose these variants of DETR as our baseline considering this efficiency.

\paragraph{Video object detection.}
Video object detection (VOD) methods aim to address the challenging cases, such as motion blur, and occlusion, suffered from the single frame. To tackle this problem, many studies have focused on improving the performance of the current frame by leveraging temporal information across videos. 
For example, FGFA~\cite{zhu2017flow}, MANET~\cite{wang2018fully}, and THP~\cite{zhu2018towards} utilize optical flow derived from FlowNet~\cite{dosovitskiy2015flownet} by aligning and aggregating the nearby features from current frames. 
TPN~\cite{kang2017object} and TCNN~\cite{kang2017t} exploit LSTM~\cite{hochreiter1997long} to construct temporal coherence between detected bounding boxes. 
To capture long-range dependencies, numerous methods adopt self-attention mechanism. Among them, SELSA~\cite{wu2019sequence} presents to use of global temporal cues by taking the full-sequence level feature aggregation. OGEMN~\cite{deng2019object} proposes to use object-guided external memory for further global aggregation.
MEGA~\cite{chen2020memory} presents a memory module that considers aggregating global and local information to enhance the feature representation. 
Recently, TransVOD~\cite{zhou2022transvod} extends the DETR detector into the video object detection domain via a temporal Transformer.

\paragraph{Test-time adaptation.}
Test-time adaptation (TTA) attempts to adapt pre-trained models to test data without relying on source domain data or incurring labeling costs. Existing TTA methods~\cite{wang2020tent,wang2022continual,li2016revisiting} typically recalibrate batch normalization (BN) layers using a batch of test samples. 
However, since Transformers typically do not contain a BN layer, it is not appropriate to apply the re-estimating BN statistics method to Transformer-based models.
Alternatively, several studies~\cite{chen2022contrastive, iwasawa2021test, jang2022test} adopt pseudo labels generated at test time for updating the model.
In the domain of object detection, TTAOD~\cite{chen2023stfar} focuses on enhancing real-time robustness across target domains via self-training and feature distribution alignment.
This paper introduces a test-time adaptation technique suitable for Transformer-based image detectors, leveraging a newly designed memory module.
\section{Methodology}
In this section, we first review DETR framework~\cite{carion2020end}, and then introduce our proposed framework, called CETR, which is a single frame context-aware object detector with context memory module (CMM), score-based sampling strategy, and memory-guided Transformer decoder (MGD) in detail.

\subsection{Preliminaries: Revisiting DETR}
DETR and its variants are based on encoder-decoder Transformer architecture. The encoder layer consists of a multi-head self-attention and feed-forward network (FFN), and the decoder layer has additional cross-attention layers. Specifically, given an input image $I$, CNN backbone extract feature map ${F}\in \mathbb{R}^{H\cdot W \times d}$, where $d$ denotes the dimension of feature and $H, W$ are the height and width of the feature map, respectively. 
Then $F$ augmented with positional encoding are fed into the Transformer encoder (denoted by $\text{Enc}(\cdot)$):
\begin{equation}
\mathcal{F} = \text{Enc}(F).
\end{equation}
Note that we omit the positional encoding in the description for clarity.
$\text{Enc}(\cdot)$ is composed of self-attention layers, which would be applied to $F$ to generate the query $Q$, key $K$, and value $V$ vectors for exchanging information features at all spatial positions. Self-attention of the Transformer encoder is conducted as:
\begin{equation}
\text{Attn}(Q=F, ~K= F, ~V=F),
\end{equation}
where multi-head attention is represented as:
\begin{equation}
\text{Attn}(Q,K,V) = \text{Softmax}(\frac{QK^T}{\sqrt{d}})V.
\end{equation}
The image feature $\mathcal{F}$ is input to the Transformer decoder, with object queries $\mathcal{O}$. 
The Transformer decoder is composed of the following two types of attention layers: multi-head self-attention and multi-head cross-attention.
\begin{equation}\mathcal{O}^{l}_{\text{sa}} = \text{Self-Attn}(Q=\mathcal{O}^{l}, ~K=\mathcal{O}^{l}, ~V=\mathcal{O}^{l}),
\end{equation}
\begin{equation}\mathcal{O}^{l+1}=\text{Cross-Attn}(Q=\mathcal{O}^{l}_{\text{sa}}, ~K=\mathcal{F}, ~V=\mathcal{F}),
\end{equation}
where decoder blocks are repeated $L$ times and $\mathcal{O}^{l}$ means object queries of $l$-th decoder block. Then, the final object queries $\mathcal{O}^{L+1}$, which have acquired semantic information from image features, are passed through a feedforward neural network (FFN) for classification and box regression. Finally, by the Hungarian algorithm, one-to-one matching between predicted objects and their corresponding ground-truth targets is established. We propose a memory module that can be applied to single frame DETR-like methods for processing video data.

\subsection{Our Approach: CETR}
\paragraph{Overview.}
Most VOD methods that utilize memory modules or reference images have a large memory footprint, which limits the amount of information that can be used at one time. To alleviate this, recent methods randomly sample memory~\cite{deng2019object, chen2020memory} or utilize information from frames close to the current frame ~\cite{zhou2022transvod, cui2023feature}. However, these approaches also have the drawback of either requiring information from the entire frame beforehand or being able to reference unnecessary information. To address these limitations, we propose CETR that selectively stores only the necessary information from the data to form a class-specific memory with a fixed size, and utilizes only the information useful for the current frame in memory.
This method enables models to efficiently leverage contextual information from the entire dataset for each frame. For this, as illustrated in Fig.~\ref{fig:main_figure}, we introduce three modules applicable to single frame DETR-like methods for the video data: 1) the context memory module (CMM), which stores contextual information of the entire dataset in a fixed size; 2) the score-based sampling, which samples only the necessary information from memory for the current frame data; and 3) the memory-guided Transformer decoder (MGD), which enhances the semantic information of object queries using the sampled spatio-temporal memory information.

\paragraph{Context Memory Module.}
Most single frame DETR-like methods employ a Transformer encoder to aggregate spatial information from current image features, enabling each image feature to include spatial context information from the current image. However, when the CNN backbone extracts ambiguous features from the current image due to challenges such as low image quality or part occlusion, they may struggle to effectively refine the image features. To mitigate this limitation, We use the fixed size memory obtained from the entire dataset to enhance each single frame image feature using temporal context information, without the need to directly use data from other frames.

Our proposed CMM has a multi-prototype class-wise memory
${M}\in \mathbb{R}^{{C}\cdot {K} \times d}$ with ${K}$ prototypes for each of the ${C}$ classes. In our pipeline, the feature map $F$ and $M$ would be fed into the Transformer encoder:
\begin{equation}
[\mathcal{F}, \mathcal{M}] = \text{Enc}([F, M]),
\end{equation}
where [$\cdot$, $\cdot$] means concatenation. 
% ${M}$ is embedded by class-wise encoding. 
In the Transformer encoder, the current information of $F$ and the spatio-temporal contextual information of $M$ are aggregated. Consequently, image feature $\mathcal{F}$ gains rich contextual information from $M$, and simultaneously, encoded memory feature $\mathcal{M}$ obtains class information fitted to the current image. Then, $\mathcal{M}$ is passed to the score-based sampling module to obtain the classification score of the current image, and $\mathcal{F}$ is forwarded to the Transformer decoder similar to DETR. 

$\mathcal{F}$ is also utilized in the context memory module for memory update. 
% \{\bar{m}_{c,k}\}_{c\leq C,k\leq K}
The context memory module extracts $N$ instance features $\tilde{\mathcal{F}}=\{f_{n}\}^N_{n=1}$ from image feature $\mathcal{F}$ and the set of class-wise memory $M=\{m_{c,k}\}^{C,K}_{c,k=1}$ is updated as:
\begin{equation}
m_{c,k_n} \leftarrow \alpha m_{c,k_n} + (1 - \alpha) f_n,~~~~
\end{equation}
\begin{equation}
\text{where} ~~k_n=\operatorname{argmax}_k\{\langle f_n, m_{c,k}\rangle \}^K_{k=1},
\end{equation}
where $\langle \cdot, \cdot \rangle$ is defined as correlation between two features, and $\alpha \in[0, 1]$ is a momentum coefficient. During training, $\tilde{\mathcal{F}}$ is extracted from the ground-truth box, while during inference, $\tilde{\mathcal{F}}$ is extracted from the predicted box. Aggregated $\mathcal{F}$ with $M$ in the Transformer encoder contains abundant contextual information. Accordingly, $M$ updated recurrently by $\mathcal{F}$ in each image of the video dataset acquires contextual information about the entire dataset that was previously observed. Additionally, the class-wise memory with multi-prototypes can also accommodate diverse distributions of instance features appearing in the entire dataset.
\begin{figure}[t]
\centering
\includegraphics[width=1.0\linewidth]{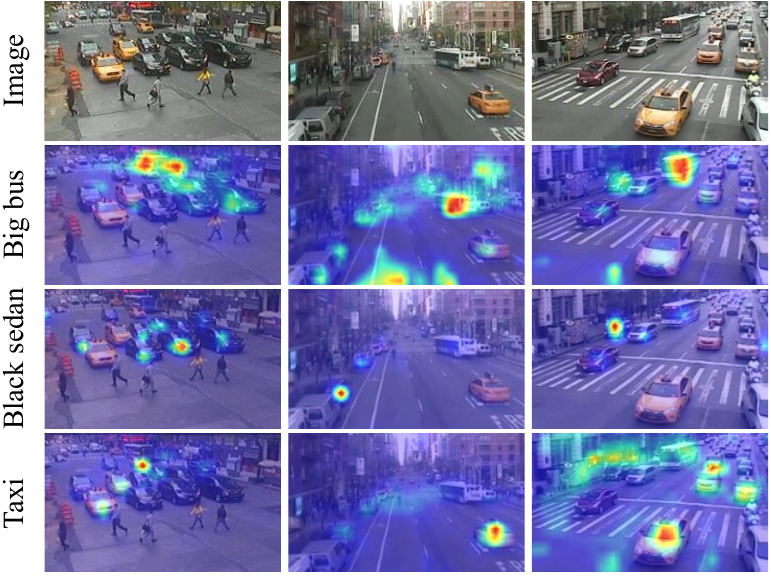}
\vspace{-10pt}
\caption{Visualization of the attention map. For certain classes, we present an attention map showing the correlation between class-wise memory and current image features. }
\label{fig:attention}
\vspace{-15pt}
\end{figure}

\begin{figure*}[t]
\begin{center}
\renewcommand{\thesubfigure}{}

\subfigure[]
{\includegraphics[width=0.24\linewidth]{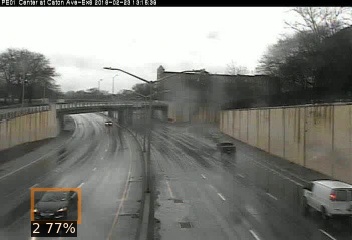}}
\subfigure[]
{\includegraphics[width=0.24\linewidth]{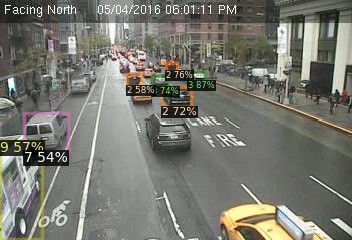}}
\subfigure[]
{\includegraphics[width=0.24\linewidth]{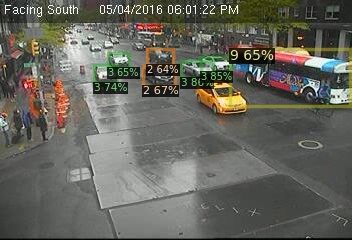}} 
\subfigure[]
{\includegraphics[width=0.24\linewidth]{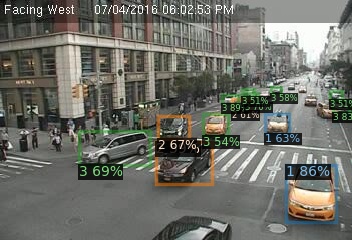}}
\hfill \\
\vspace{-20pt}

\subfigure[]
{\includegraphics[width=0.24\linewidth]{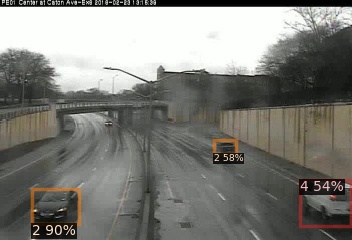}}
\subfigure[]
{\includegraphics[width=0.24\linewidth]{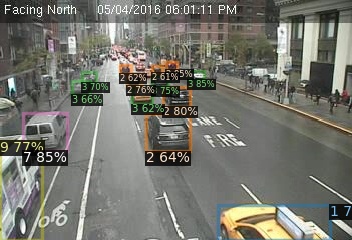}}
\subfigure[]
{\includegraphics[width=0.24\linewidth]{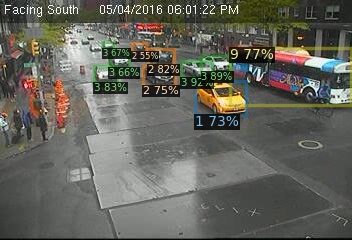}}
\subfigure[]
{\includegraphics[width=0.24\linewidth]{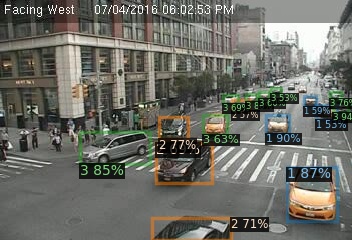}}
\hfill \\
\vspace{-10pt}

\end{center}
\vspace{-16pt}
\caption{Qualitative Results on the CityCam dataset~\cite{zhang2017understanding}. Comparison between the baseline~\cite{liu2022dab} (top) and our proposed method (bottom) is shown. As exemplified, our method provides more robust detection results compared to the baseline. }%
\label{fig:main_qual}\vspace{-10pt}
\end{figure*}

\paragraph{Score-based Sampling.}
Carefully selecting the relevant information from the memory is equally important as creating a high-quality memory. However, recent VOD works that employ memory modules often fail to guarantee optimal memory sampling by either randomly sampling memory or utilizing only the memory information around the current frame. Empirically, we have discovered that utilizing class information from the current image to selectively sample memory leads to significant performance improvement. Detailed results of the related experiments are provided in Table~\ref{tab:sampling}. Motivated by this, we introduce a score-based sampling module to extract information relevant to the current image from the class-wise memory $M$ (see Fig.~\ref{fig:sampling}). The score-based sampling module consists of two parts: a classification part and a multi-threshold sampling part. In the classification part, the sampling module obtains classification score $p_{c,k}\in [0, 1]$ by passing encoded memory feature $\mathcal{M}=\{\mathfrak{m}_{c,k}\}^{C,K}_{c,k=1}$, which contains the class information of the current image. This operation is executed via a classification head $\text{FFN}_{c,k}$ composed independently for each $\mathfrak{m}_{c,k}$, followed by a sigmoid function:
\begin{equation}
p_{c,k} = \text{Sigmoid}(\text{FFN}_{c,k}(\mathfrak{m}_{c,k})).
\end{equation}
Subsequently, in the multi-threshold sampling part, the sampled memory $\tilde{M}=\{\tilde{m}_{c,k}\}^{C,K}_{c,k=1}$ is obtained by: 
\begin{equation}
\tilde{m}_{c,k} = \text{Proj}(\tilde{m}^1_{c,k},
\tilde{m}^2_{c,k}, \ldots, \tilde{m}^T_{c,k}), \\
\end{equation}
\begin{equation}
\text{where} ~~\tilde{m}^t_{c,k} = {s^t_{c,k}} m_{c,k} + (1 - {s^t_{c,k}}) \varnothing,
\end{equation}
where $T$ is the number of sampling index ${s^{t}_{c,k}}$, and $\varnothing $ denotes learnable no-class embedding.
Here ${s^t_{c,k}}$ is derived by binarizing $p_{c,k}$ using $T$ thresholds $\tau_t$ ($i.e., \, s^t_{c,k}=\delta(p_{c,k}>\tau_t)$). The delta function $\delta$ outputs 1 when the condition is true, and 0 otherwise. The projection layer $\text{Proj}(\cdot)$ combines multi-thresholded memory information with varying confidences to generate the final sampled memory. 

During training, we employ asymmetric loss~\cite{ben2020asymmetric} additionally to train the classification head and enhance the class discrimination capability of the Transformer encoder.

\paragraph{Memory-guided Transformer Decoder.}
Many recent works that have built upon DETR-like methods address the ambiguity in the role of object queries by incorporating positional information into the object queries~\cite{meng2021conditional, liu2022dab}. This has clarified the positional information of object queries, enabling them to locate objects at various positions within the current image. However, if object queries are aggregated with poor image features of the current data, they might acquire incorrect semantic information. To address this issue, we propose a method to enhance the semantic information of object queries, using a memory-attention layer. 
One block of our proposed memory-guided Transformer decoder (MGD) is composed of three types of attention layers, formed by adding a memory cross-attention layer to the components of the existing decoder blocks:
\begin{equation}
\mathcal{O}^{l}_\text{sa} = \text{Self-Attn} (Q=\mathcal{O}^{l},K=\mathcal{O}^{l},V=\mathcal{O}^{l}),
\end{equation}
\begin{equation} \mathcal{O}^{l}_\text{ca} = \text{Cross-Attn}(Q=\mathcal{O}^{l}_\text{sa},K=\mathcal{F},V=\mathcal{F}),
\end{equation}
\begin{equation}
\begin{aligned}
\mathcal{O}^{l+1} = \text{Mem.Cross-Attn}(Q=\mathcal{O}^{l}_\text{ca},K=\tilde{M},V=\tilde{M}),
\end{aligned}
\end{equation}
where $\mathcal{O}$ means object queries. In the MGD, $\mathcal{O}$ acquires semantic information about the current image from the existing attention layers, and then enhances its associated class information through the memory cross-attention layer. Finally, each output object query of MGD is transformed by an FFN to output a class score and box location for each object. The subsequent processes, such as Hungarian matching and losses, follow DETR~\cite{carion2020end}.

\subsection{Test-time Memory Adaptation}
Given the non-parametric design of our CMM, adapting our memory to test data becomes achievable without the need for additional fine-tuning. To facilitate this, we introduce a test-time adaptation strategy utilizing CMM. Within this framework, the CMM preserves representations acquired during training and independently stores contextual information specific to the target domain.
To mitigate the storage of initial noisy representations, uniform coefficients are employed for the momentum update of individual memories during the training stage. 
During testing, on the other hand, we update the test memory module $M'$ by individually adjusting the update rate of each memory $m'_{c,k}$ to align with the memory distribution in the test domain. 
\begin{equation}
m'_{c,k} \leftarrow \frac{1}{i_{c,k}+ J}\,( i_{c,k}m'_{c,k} \, + \, \sum^{J}_{j=1}f_{j}),\\
\end{equation}
where $i_{c,k}$ indicates the total number of detected instances before current frame, and $J$ is the number of instance features $f_j$ at current frame. 
The update process involves adding to memory the instance features in the current image that are highly correlated with the mean values in test and source memory. 
For memory retrieval, a weighted sum is applied to combine memory from training and memory originating from the target domain:
\begin{equation}
M' \leftarrow \beta M + (1-\beta) M',
\end{equation}
where $\beta \in [0, 1]$ denotes the weight of the source domain.
This adaptation technique ensures that memory is adapted to the target domain while preserving important contextual information from the source distribution. 
We can also tailor the memory to specific individual cameras, resulting in a more robust test memory module for various camera systems.

\begin{table*}[t!]
	\centering
        \scalebox{1.0}{
	{
		\setlength{\tabcolsep}{9pt}
		\begin{tabular}{l|c|c|ccccc}
			\toprule
			Model & FPS $\uparrow$ & \thead{Mem $\downarrow$ \\ (GB)}  & AP & AP$_{50}$  & AP$_\text{S}$ & AP$_\text{M}$ & AP$_\text{L}$ \\ 
			\midrule
                Faster R-CNN~\cite{girshick2015fast} & 37.8& 0.42 & 23.3 & 39.6  & 18.5 & 39.0 & 36.3   \\ 
                % Context R-CNN~\cite{beery2020context} & - & - & - & $42.6$ & - & - & - & - \\ 
                \midrule
                % DETR~\cite{carion2020end} & - & - & - & - & - & - & - & -   \\
                Conditional DETR~\cite{meng2021conditional} & 36.7 & 0.46 & 23.0 & 41.9  & 17.7 & 38.9 & 43.2   \\
                Conditional DETR + CETR  & 33.8 & 0.48 & 24.4 & 42.4  & 19.3 &  40.0 & 45.0 \\
                \midrule
                
                DAB-DETR~\cite{liu2022dab} & 33.5 & 0.47 &  23.8  & 40.2 & 18.6 & 39.4 & 43.4  \\
                DAB-DETR + CETR  & 30.7 & 0.48 & 25.0 &  43.0  & 19.7 &  40.4 & 48.4 \\
                \midrule
                
                Deformable DETR~\cite{zhu2020deformable} & 37.9 & 0.42 &  24.2 & 41.0  & 20.1 & 41.4 & 48.0   \\
                Deformable DETR + TransVOD~\cite{zhou2022transvod} & 4.3  & 4.19 &  23.6 &  40.2 & 19.8 & 40.2 & 47.4   \\
                Deformable DETR + CETR  & 30.6 & 0.48 & \textbf{25.7} &  \textbf{43.0}  & \textbf{20.5} &  \textbf{41.8} & \textbf{53.6} \\
                \bottomrule
		\end{tabular}
        }
	}\vspace{-5pt}
    \caption{Quantitative results on the CityCam dataset~\cite{zhang2017understanding}.  }
    \label{tab:main}\vspace{-10pt}
\end{table*}

\section{Experiments}
\subsection{Experimental Setup}

\paragraph{Datasets.}
In this study, we assess the effectiveness of our framework through experiments conducted on two distinct datasets: the CityCam dataset~\cite{zhang2017understanding} and the ImageNet VID dataset~\cite{russakovsky2015imagenet}.
The CityCam dataset consists of approximately 60K labeled frames, with 900K annotated objects across 10 vehicle classes. It is composed of 16 camera locations in downtown and parkway areas, spanning four typical weather conditions and time periods. For our experiments, We use 13 camera locations for training and 3 camera locations for testing. 
ImageNet VID consists of 3,862 training videos and 555 validation videos across 30 object classes. Following common settings in previous works~\cite{zhou2022transvod,chen2020memory}, we train CETR on the training split of ImageNet VID and DET datasets. 

\paragraph{Implementation details.}
Our framework is trained on 24GB RTX-3090 GPUs with a batch size of 16, using the AdamW~\cite{loshchilov2017decoupled} optimizer. We train CETR for 150K iterations, with a learning rate of ${10^{-4}}$ for the first 120K iterations and ${10^{-5}}$ for the last 30K iterations.
For fast convergence, we employed variants of DETR as our baseline.  In the CityCam experimentation, ResNet-50~\cite{he2016deep} is used as the backbone and initialized with pre-trained weights from ImageNet dataset~\cite{russakovsky2015imagenet}, while the 
Transformer encoder and decoder were initialized randomly.
In the ImageNet VID experiment, ResNet-101 is used as the backbone and  
the entire network is initialized with pre-trained weights from COCO dataset~\cite{lin2014microsoft}. For the ImageNet VID experiment and the ablation study, we used DAB-DETR with CETR.

\paragraph{Evaluation metric.}
We follow the standard COCO evaluation. We report the average precision under different IoU thresholds (AP), AP scores at IoU thresholds are 0.5 (AP$_{50}$), and different object scales (AP$_\text{S}$, AP$_\text{M}$, AP$_\text{L}$).
For the ImageNet VID dataset, we follow the common protocol~\cite{zhou2022transvod,chen2020memory,deng2019relation} and leverage average precision at IoU thresholds are 0.5 (AP$_{50}$) as the evaluation metric.

\subsection{Experimental Results}
\begin{table}[t]\centering

\scalebox{0.84}{%
\begin{tabular}{l|c|c|ccc}
\toprule
Model   & Online &  ~~AP$_{50}$~~ &  FPS $\uparrow$ & \thead{\#Params $\downarrow$ \\ (M)} & \thead{Mem $\downarrow$ \\ (GB)}   \\
\midrule

% THP  & 78.6 & 13.0 &-&-    & Res101+DCN \\

SELSA  & $\tikzxmark$  & 80.3  & 7.2  &- &-      \\
LRTR  & $\tikzxmark$ & 80.6  & 10 &- &-     \\

RDN   & $\tikzxmark$ & \textbf{81.8}  &  10.6 &- &-     \\
%PSLA  & 80.0 & 13.3 & 72.2 &-     & Res101+DCN  \\

%LSTS  & 80.1 & 21.2 & 65.5  &-    & Res101+DCN  \\

TransVOD Lite   & $\tikzxmark$  &80.5 & {32.3} &74.2& 2.94    \\
\midrule
DFF  & $\tikzcmark$  & 73.1 & 20.25 &97.8   &-     \\
D\&T   & $\tikzcmark$& 75.8 & 7.8 &- &-      \\
LWDN   &$\tikzcmark$  & 76.3 & 20  &77.5 &-   \\
OGEMN &   $\tikzcmark$ & 76.8  & 14.9 &- &-    \\
PSLA   & $\tikzcmark$&77.1 & 18.7 & {63.7}&-       \\
LSTS   &   $\tikzcmark$ &77.2 & 23.0 & 64.5&-      \\

\midrule
\textbf{CETR} &$\tikzcmark$ & \textbf{79.6} & {23.3} & 65.7& 0.55   \\

%\hline
\bottomrule
\end{tabular}
}
\caption{
Performance comparison with state-of-the-art real-time VOD methods with ResNet-101 backbone on ImageNet VID dataset~\cite{russakovsky2015imagenet}. Here we use AP$_{50}$, which is commonly used as mean average precision (mAP) in other VOD methods.
}
\label{table:imagenetvid}
\vspace{-15pt}
\end{table}

\paragraph{Quantitative results.}
Table~\ref{tab:main} presents our main results on the CityCam testing set, which we have divided. We applied our proposed framework to the single frame DETR-like methods and conducted quantitative comparisons with other single frame detection methods and a multi-frame DETR-like method, TransVOD~\cite{zhou2022transvod} that use Deformable DETR as their baseline. Compared to the single frame baseline~\cite{zhu2020deformable}, our method showed improvements of 1.5\% AP and 2.0\% AP$_{50}$, with only a marginal increase of 0.06 GB in allocated memory and a decrease of 7.3 FPS, while the multi-frame DETR-like method showed an increase of 3.77 GB in allocated memory and a decrease of 33.6 FPS. Additionally, the multi-frame method that has mainly been utilized with video clip data of consistent short-frame intervals exhibits poor performance on the CityCam dataset with wider frame intervals.
We also compare our approach with SELSA~\cite{wu2019sequence}, LRTR~\cite{shvets2019leveraging}, RDN~\cite{deng2019relation}, TransVOD Lite~\cite{zhou2022transvod}, DFF~\cite{zhu2017deep}, D\&T~\cite{feichtenhofer2017detect}, LWDN~\cite{jiang2019video}, OGEMN~\cite{deng2019object}, PSLA~\cite{guo2019progressive}, and LSTS~\cite{jiang2020learning} on ImageNet VID dataset. As shown in Table~\ref{table:imagenetvid}, our method demonstrates competitive performance with other VOD methods without the need for multi-frame approaches, which require significant allocated memory and disrupt general online inference. 
Among online methods, it achieves the highest AP$_{50}$ of 79.6\%. Furthermore, our method exhibits higher FPS compared to most approaches, except for the method that requires multi-frame image inference at once.

\paragraph{Qualitative results.}
As shown in Fig.~\ref{fig:attention}, we provide a visualization of the correlation between the proposed memory and image features. To do this, we utilize the attention map in the 4th layer of the Transform encoder. The CMM exhibits notable attention scores for objects of the corresponding class. This observation emphasizes that class-specific memory clearly contains the relevant information specific to each class.
In addition, we visually compare the object detection results of the baseline model and our approach in Fig.~\ref{fig:main_qual}. 
The results showcase higher confidence scores for most classes in our approach compared to the baseline. Notably, comparing the 2nd and 4th columns, we notice that our model performs well in detecting objects that are partially visible within the frame, while the baseline fails. 
In addition, the first qualitative result shows the superiority of our model in capturing even rare classes (e.g., small trucks) within the dataset.

\begin{table}[t]
\centering
\resizebox{0.7\linewidth}{!}{
\begin{tabular}{l|cccc|cc} 
\toprule
Methods & CMM & MGD & SS & MT   & AP$_{50}$   \\ 
\midrule
Baseline    & -      & -          & -       & -                   & 40.2 \\ 
\midrule
\multirow{4}{*}{Ours}   & $\checkmark$  & -      & -    & -    & 40.5   \\
                        &$\checkmark$   & $\checkmark$  & - &  -             &  40.5 \\
                        &$\checkmark$  & $\checkmark$& $\checkmark$  & -   & 41.1  \\ 
                        &$\checkmark$ &$\checkmark$  & $\checkmark$  & $\checkmark$   & \textbf{42.5}   \\ 
\bottomrule
\end{tabular}
}
\vspace{-5pt}
\caption{Ablation study on main components. CMM, SS, MT, and MGD denote Context Memory Module, score-based sampling, Multi-level thresholding, and Memory-Guided Transformer Decoder, respectively.}\label{tab:ablation}
\vspace{-5pt}
\end{table}

\begin{table}[t!]
	\centering
	\scalebox{1.0}{
		% \small
		\setlength{\tabcolsep}{0.5em}
		\begin{tabular}{c|ccccc}
		\toprule
			
        \# Prototype ($K$) & AP  & AP$_{50}$   & AP$_\text{S}$  & AP$_\text{M}$  & AP$_\text{L}$  \\
        \midrule
        1 & 23.7 & 41.9 & 18.2 & 38.9 & 46.6\\
        3 & \textbf{24.8} & \textbf{42.5}  & \textbf{19.6}& \textbf{40.0}& \textbf{47.4} \\
        5 & 23.2 & 41.1& 17.4& 38.8& 45.1\\
        10 & 23.3 & 40.0& 18.2& 39.1& 45.6\\
        \bottomrule
	    \end{tabular}
 	}
    \vspace{-5pt}
    \caption{Performance for the number of prototypes per class.}
    \label{tab:ppc}
    \vspace{-10pt}
\end{table}

\subsection{Ablation Study and Analysis}

\paragraph{Memory module analysis.}
We first investigate the effect of our main components. As shown in Table~\ref{tab:ablation}, when our CMM is used only in the Transformer encoder, it has a 0.3\% improvement AP$_{50}$ compared to the single frame baseline~\cite{liu2022dab}. When used in isolation, the MGD does not exhibit any performance enhancement. However, when used in combination with the score-based sampling method, there was an increase of 0.6\% AP$_{50}$. In addition, when used in conjunction with the multi-level thresholding method, it shows an additional improvement of 1.4\% AP$_{50}$.

\begin{table}[t]
    \footnotesize
    \centering{
    \begin{tabular}{l|c} 
    \toprule
    Method   & AP$_{50}$    \\ 
    \midrule
    Baseline     & 40.2   \\
    \midrule
    Learnable memory     & 40.9   \\
    Full memory  & 40.5    \\
    Random sampling      & 41.2  \\ 
    Score-based sampling    & \textbf{42.5} \\
    \midrule
    GT sampling (oracle)        & 57.6 
    \\
    \bottomrule
    \end{tabular}
    \vspace{-5pt}
    
    }
        \caption{Experiments on sampling strategy.
    }
    \label{tab:sampling}
    \vspace{-5pt}
\end{table}

\paragraph{The number of prototypes.}
Table~\ref{tab:ppc} illustrates the ablation study on the number of prototypes of each class in our context memory module. When using one prototype and three prototypes, our approach achieves performance improvements of 1.7\% and 2.3\% AP$_{50}$ compared to the single frame baseline, respectively. However, we observe that the performance of our approach in the CityCam dataset decreases as the number of prototypes is increased beyond 3. This outcome is believed to be due to the CityCam dataset consisting solely of classes grouped under the vehicle category that share similar types. As a result, a small number of prototypes is enough to represent the distribution of class features, and too many prototypes may, in fact, hinder the utilization of class features.

\begin{table}[t]
    \footnotesize
    \centering{
    \begin{tabular}{l|ccccc} 
    \toprule
        Method & AP  & AP$_{50}$   & AP$_\text{S}$  & AP$_\text{M}$  & AP$_\text{L}$  \\
        \midrule
        Our Baseline  & 24.8 & 42.5 & 19.6 & 40.0 & 47.4\\
        \midrule
        + Memory update  & 24.9 & 42.8 & \textbf{19.7} & 40.2 & 47.8\\
        + Cam specific & \textbf{25.0} & \textbf{43.0}  & \textbf{19.7}& \textbf{40.4}& \textbf{48.4} \\
        \bottomrule
    \end{tabular}
    \vspace{-5pt}
    }
        \caption{Experiments on the test-time memory adaptation.
    }
    \label{tab:tta}
    \vspace{-15pt}
\end{table}

\paragraph{Sampling strategy.}
Table~\ref{tab:sampling} reports the performance of our approach according to various class-wise memory sampling strategies for the memory-guided Transformer decoder. When employing the class-wise memory sampling strategy using ground-truth images, If we have an oracle-level knowledge of the correct answers, it shows a significant performance improvement of 17.4\% AP$_{50}$. Taking this result as motivation, we designed our score-based sampling module. The performance of the approaches improves in each case when using all learnable memory or randomly sampling class-wise memory. However, the classification score-based sampling strategy leads to the most improvement in performance. This result demonstrates the effectiveness of our score-based sampling method.

\paragraph{Test-time memory adaptation.}
Lastly, we conduct extensive experiments to assess the effectiveness of the CMM-based test-time adaptation approach, as shown in Table~\ref{tab:tta}. 
The results highlight that the memory update technique adapted to the target domain yields meaningful performance improvements without requiring further training or parameter optimization. From the table, we also notice that by adding a camera-specific way of configuring memory, performance can be improved by 0.5\% for the $AP_{50}$ over the baseline.

\section{Conclusion}

We proposed a new framework called CETR applicable to single frame DETR-like methods for video object detection. To handle video data with a single frame approach, we introduced a context memory module that enables the use of spatio-temporal contextual information from the entire dataset. In addition, we used a score-based sampling and a memory-guided transformer decoder to effectively make use of our context memory. Our method exhibited a meaningful performance improvement over the single frame baseline in the CityCam dataset, with only a slight increase in allocated memory and a low decrease in FPS. Furthermore, our method demonstrated a remarkable performance improvement over other real-time online video object detection methods when evaluated on the ImageNet VID dataset.

\section{Acknowledgments}
This research was supported by the MSIT, Korea (IITP-2023-2020-0-01819, RS-2023-00222280), National Research Foundation of Korea (NRF-2021R1C1C1006897, NRF-2018M3E3A1057288).

\bibliography{aaai24}

\clearpage
\renewcommand{\thesection}{\Alph{section}}
\setcounter{section}{0}
\section*{Appendix}
The following sections present more examination of our results, including a detailed explanation of our methodology, as well as supplementary visualizations of our sampling outcomes. Section~\ref{sec:imp} delves into the specific implementation details of our method, including used losses, hyperparameter settings, and data augmentations of our approach. In Section~\ref{sec:pseudo}, we offer a pseudocode of our context memory module's update strategy and TTA method. Finally, Section~\ref{sec:qual} shows additional qualitative results of CETR in CityCam dataset and ImageNet VID dataset.

\section{A. Implementation Details}\setlabel{A}{sec:imp}
\subsection{A.1 Architecture}
\begin{figure*}[ht]
\begin{center}
\includegraphics[width=1.0\linewidth]{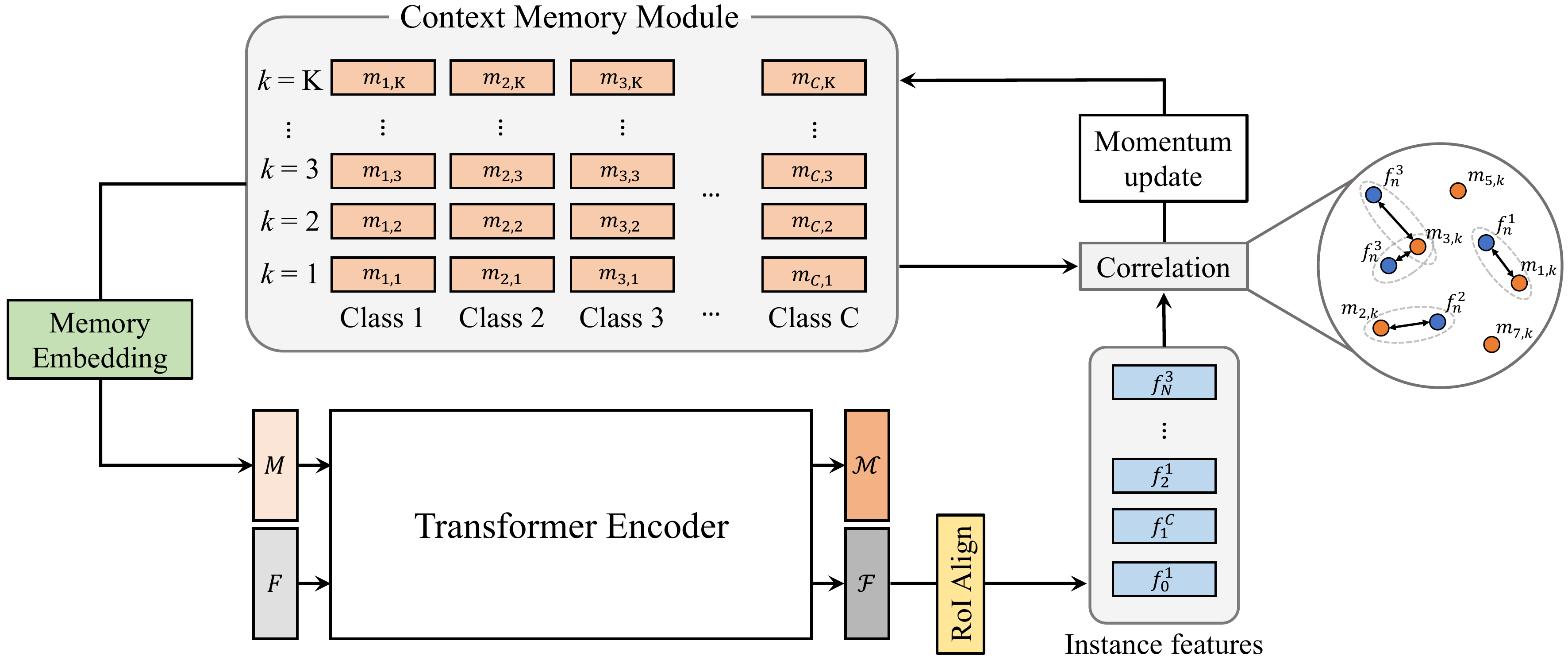}
\end{center}
\vspace{-10pt}
\caption{\textbf{Details of context memory module update.} }
\label{fig:memory_detail}
\vspace{-5pt}
\end{figure*}

\begin{figure*}[ht!]
\centering
\includegraphics[width=0.5\linewidth]{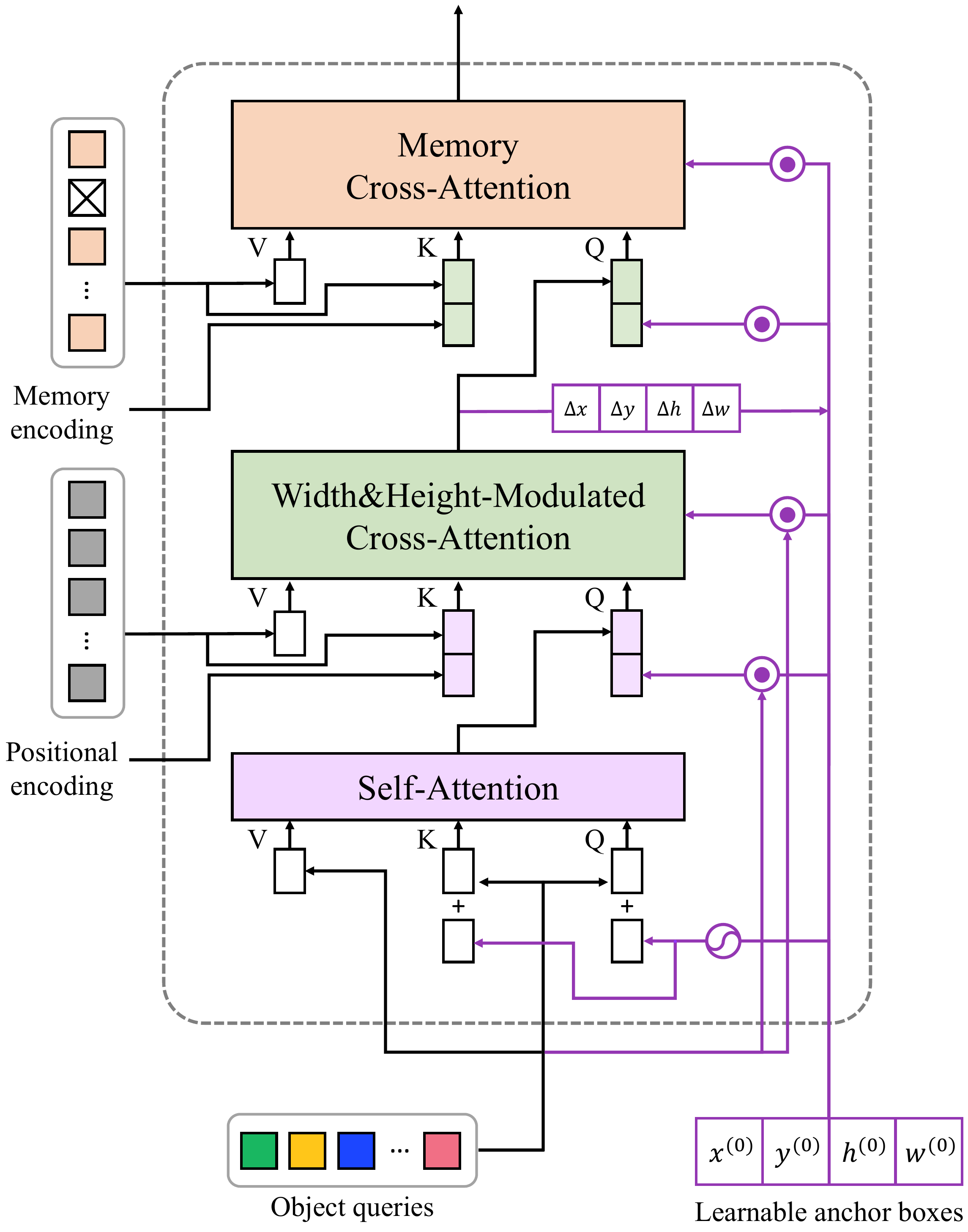}
% \vspace{-15pt}
\caption{\textbf{Detail illustrations of our memory-guided Transformer decoder (MGD)}  }
\label{fig:decoder_detail}
\vspace{-15pt}
\end{figure*}

We employ DAB-DETR which includes a CNN backbone, a Transformer with modified decoder architecture, and prediction heads for box regression as our single-frame baseline. Additionally, we introduce the context memory module that has a multi-prototype class-wise memory. As illustrated in Fig.~\ref{fig:memory_detail}, the class-wise memory is updated by instance features extracted from the image feature. Each instance feature with an assigned class is updated to the prototype of the corresponding class memory that has the highest correlation score using the momentum update strategy. To clarify the role of object queries, DAB-DETR introduced the learnable anchor boxes, which provide explicit positional information to the object queries. Furthermore, we leverage the memory module to enhance the semantic information of object queries (see Fig.~\ref{fig:decoder_detail}). We use ImageNet-pretrained ResNet as our backbone. In the experiments on ImageNet VID, we use ResNet-101 and on CityCam, we use ResNet-50. Following DAB-DETR, we utilize 6 encoder blocks, 6 decoder blocks, and 300 object queries.

\subsection{A.2 Training Details}
In this subsection, we present the details of the hyper-parameters in our model and training setup. We set the hyper-parameters: $\alpha$ = 0.99, $\beta$ = 0.6.  Additionally, we used $K$ = 10 for ImageNet VID and $K$ = 3 for CityCam. In the experiments on ImageNet VID, we employ a pre-trained DAB-DETR on COCO as our single-frame baseline, while for the experiments on CityCam, we trained the model from scratch. We use the same data augmentation as adopted in DETR during training: resizing, random cropping, and random horizontal flipping. For scale augmentation, we resize the input images to ensure that the shortest side falls within the range of 480 to 800 pixels, while the longest side remains under 1333 pixels. To enhance the learning of global relationships through encoder self-attention, we employ random crop augmentations during training. Here, each image is subjected to a 0.5 probability of being cropped into a random rectangular patch, which is subsequently resized to dimensions between 800 and 1333.

\subsection{A.3 Dataset}
The CityCam dataset~\cite{zhang2017understanding} is a collection of fixed traffic videos from each camera at four hourly intervals each day (7am-8am, 12pm-1pm, 3pm-4pm, and 6pm-7pm). These cameras have a frame rate of about 1 fps with a resolution of 352×240. With these properties, the CityCam dataset has a low correlation between neighboring frames because each clip has a fixed background and long intervals between frames. The dataset has 10 different vehicle types as object classes, including taxis, black sedans, other cars, small trucks, medium trucks, large trucks, vans, medium buses, large buses, and other vehicles. Due to this class categorization, there is a high degree of similarity between the different classes. 
We conduct experiments to demonstrate that CETR performs robustly on these datasets, which are rarely used in VOD methods but are very essential in the real world.
On the other hand, the ImageNet VID  dataset~\cite{russakovsky2015imagenet} is a large-scale public dataset for video object detection. We experiment on this dataset to compare its performance with other VOD approaches and to show the efficacy of CETR on common video data. 

\subsection{A.4 Loss Details}
Following DAB-DETR, we use focal loss~\cite{lin2017focal} for classification loss  and L1 loss and GIOU loss~\cite{rezatofighi2019generalized} for bounding box loss. The same losses are used for calculating the matching cost of Hungarian algorithm and final training loss calculating, while they have different weights. Classification loss have weight of 2.0 when used in Hungarian algorithm and have weight of 1.0 when used in the final loss. L1 loss and GIOU losses have the weight of 5.0 and 2.0 in both cases. To train our score-based sampling module, we additionally used Asymmetric loss~\cite{ridnik2021asymmetric} for multi-label classification. Asymmetric loss only used for final loss calculation with weight of 0.005. 

\section{B. Pseudo Code of Memory Update}\setlabel{B}{sec:pseudo}
We provide detailed algorithm of memory update strategy and test-time memory adaptation.
See Algorithm~\ref{alg:Training} for training-time memory update and Algorithm~\ref{alg:TTA} for test-time memory update.

\section{C. Additional Qualitative Results}
\begin{figure*}[ht!]
\begin{center}
\renewcommand{\thesubfigure}{}

\subfigure[]
{\includegraphics[width=0.4\linewidth]{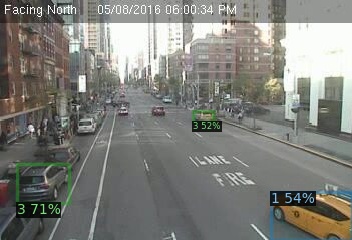}}
\subfigure[]
{\includegraphics[width=0.4\linewidth]{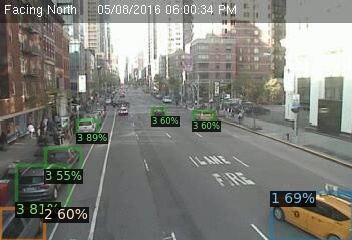}}
\hfill \\
\vspace{-20pt}

\subfigure[]
{\includegraphics[width=0.4\linewidth]{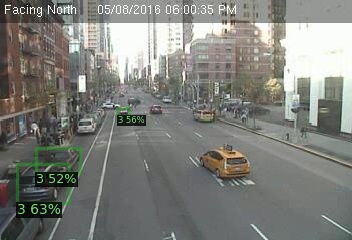}}
\subfigure[]
{\includegraphics[width=0.4\linewidth]{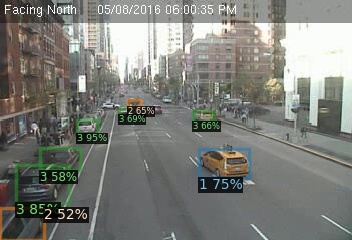}}
\hfill \\
\vspace{-20pt}

\subfigure[]
{\includegraphics[width=0.4\linewidth]{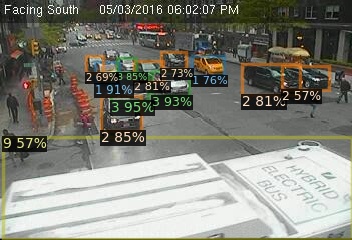}}
\subfigure[]
{\includegraphics[width=0.4\linewidth]{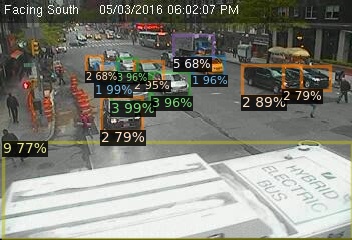}}
\hfill \\
\vspace{-20pt}

\subfigure[]
{\includegraphics[width=0.4\linewidth]{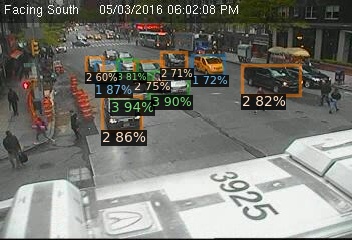}}
\subfigure[]
{\includegraphics[width=0.4\linewidth]{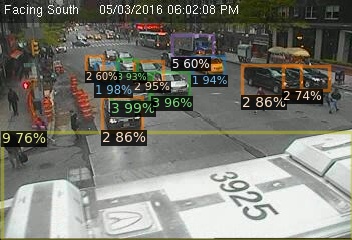}}
\hfill \\
\vspace{-10pt}

\end{center}
\vspace{-16pt}
\caption{\textbf{Qualitative Results on CityCam dataset~\cite{zhang2017understanding}} comparison between the baseline~\cite{liu2022dab} (left) and our proposed method (right). As exemplified, our method provides more robust detection results when compared to the baseline. }%
\label{fig:citycam}\vspace{-10pt}
\end{figure*}

\begin{figure*}[ht!]
\begin{center}
\renewcommand{\thesubfigure}{}

\subfigure[]
{\includegraphics[width=0.41\linewidth]{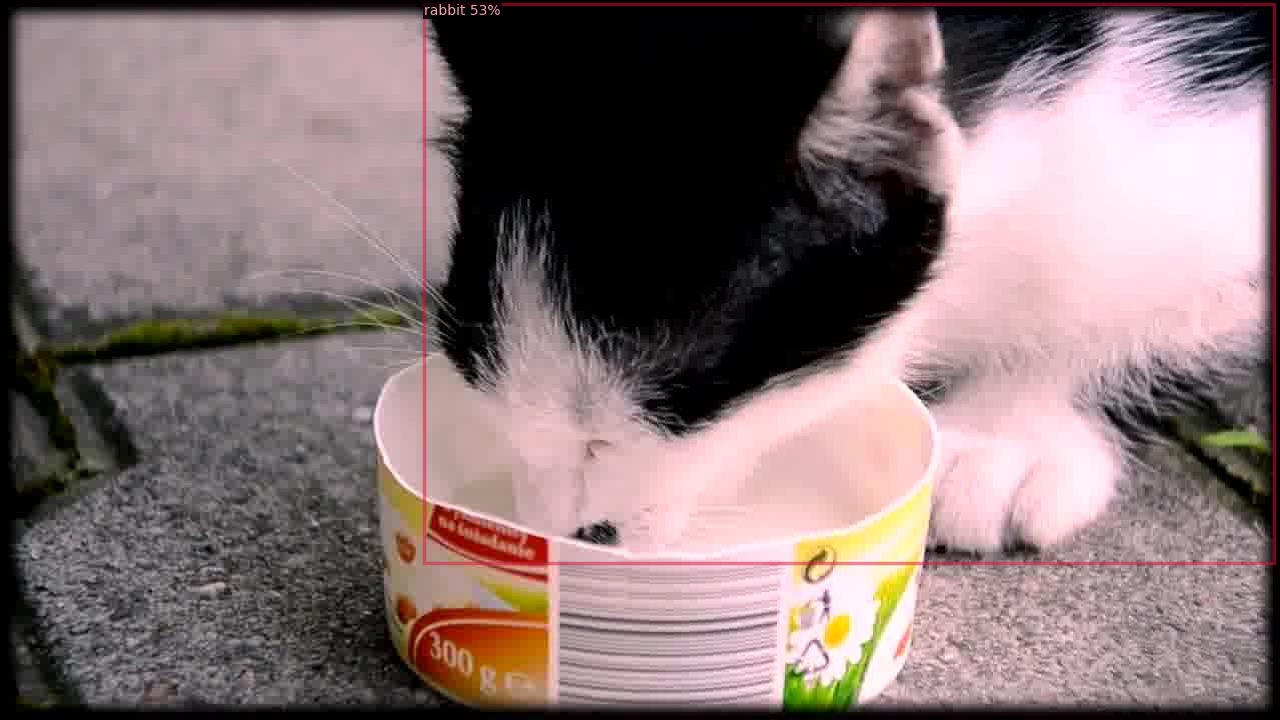}}
\subfigure[]
{\includegraphics[width=0.41\linewidth]{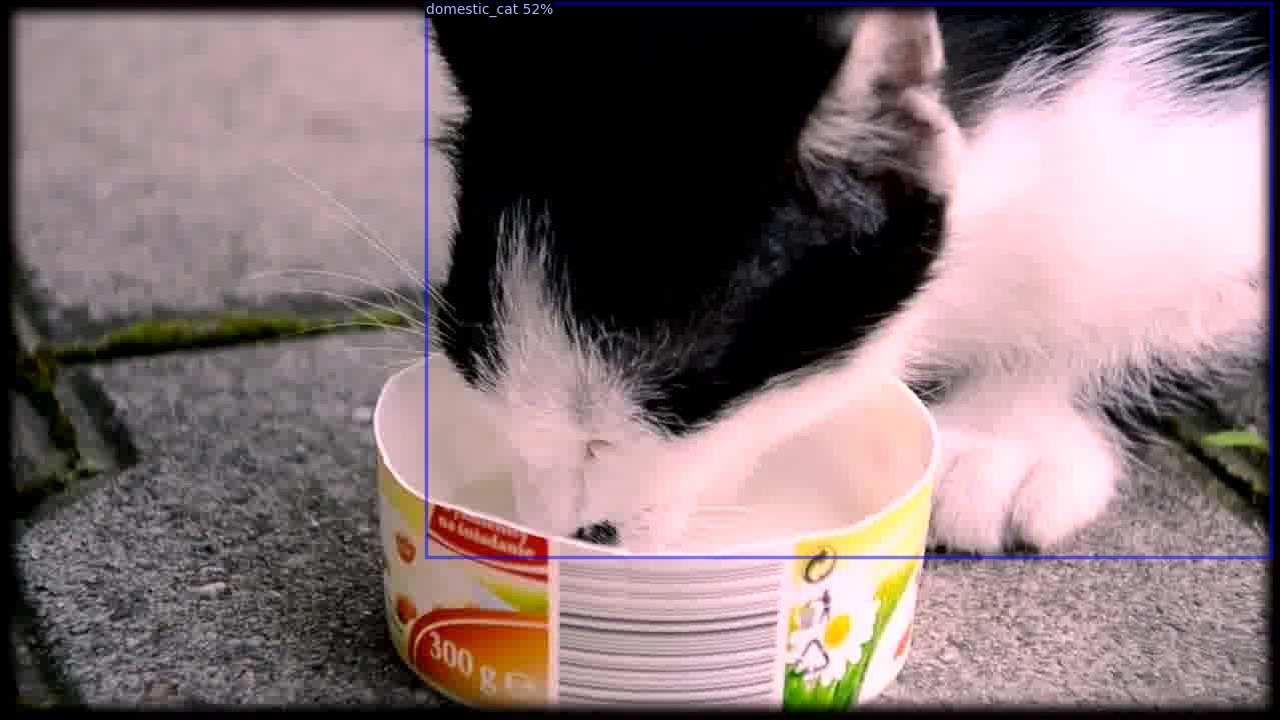}}
% \subfigure[]
% {\includegraphics[width=0.418\linewidth]{Figures/supp/vid_qual/base/000029.JPEG}}
% \subfigure[]
% {\includegraphics[width=0.418\linewidth]{Figures/supp/vid_qual/base/000029.JPEG}}
\hfill \\
\vspace{-20pt}

\subfigure[]
{\includegraphics[width=0.41\linewidth]{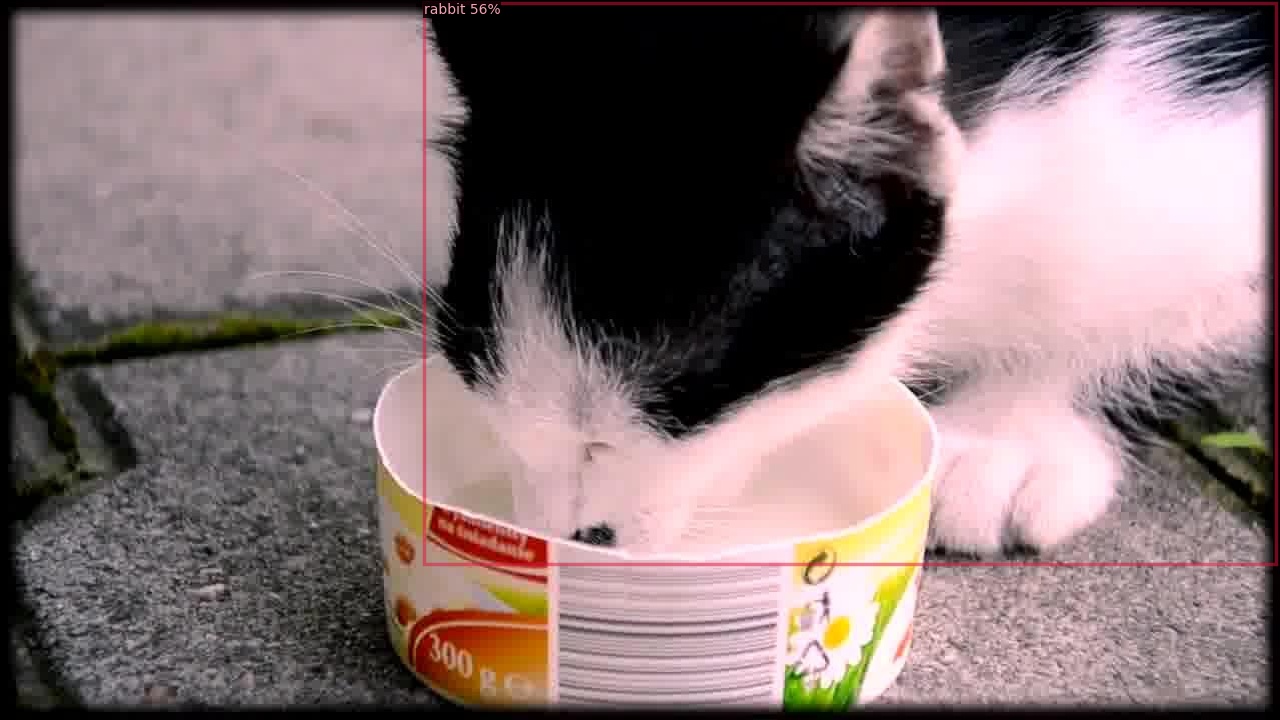}}
\subfigure[]
{\includegraphics[width=0.41\linewidth]{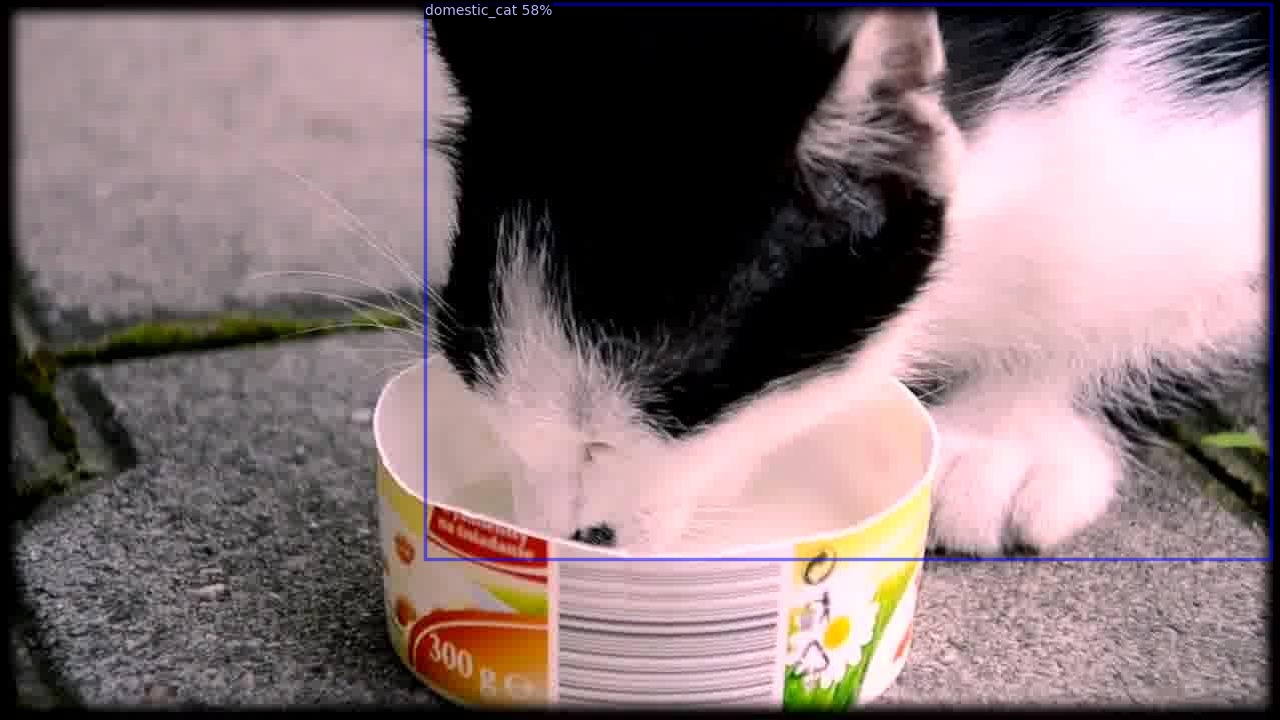}}
% \subfigure[]
% {\includegraphics[width=0.418\linewidth]{Figures/supp/vid_qual/base/000029.JPEG}}
% \subfigure[]
% {\includegraphics[width=0.418\linewidth]{Figures/supp/vid_qual/base/000029.JPEG}}
\hfill \\
\vspace{-20pt}

\subfigure[]
{\includegraphics[width=0.41\linewidth]{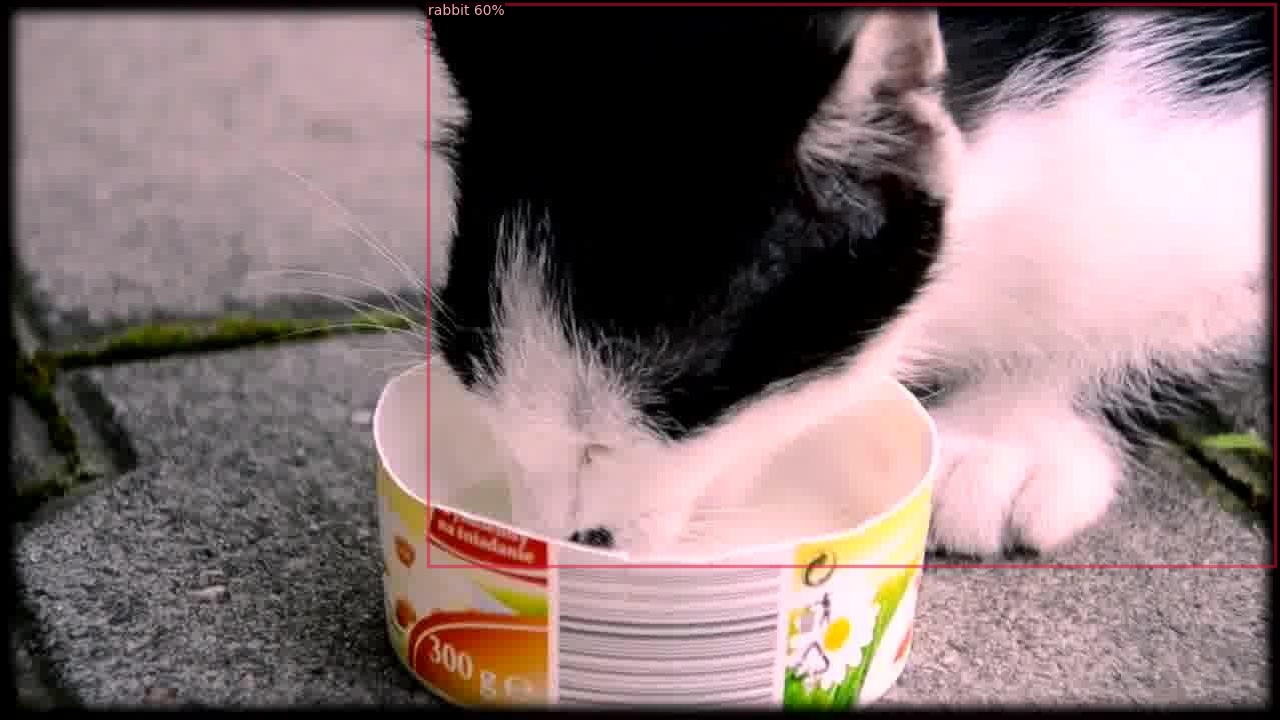}}
\subfigure[]
{\includegraphics[width=0.41\linewidth]{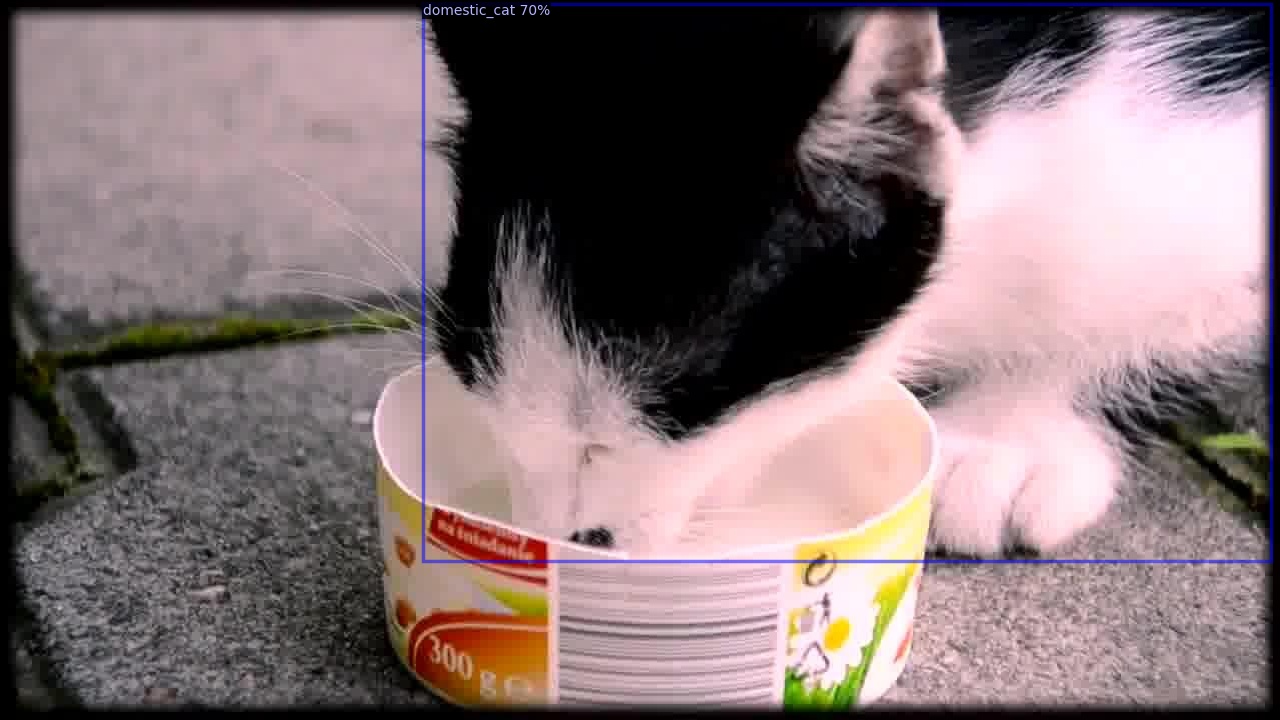}}
% \subfigure[]
% {\includegraphics[width=0.418\linewidth]{Figures/supp/vid_qual/base/000029.JPEG}}
% \subfigure[]
% {\includegraphics[width=0.418\linewidth]{Figures/supp/vid_qual/base/000029.JPEG}}
\hfill \\
\vspace{-20pt}

\subfigure[]
{\includegraphics[width=0.41\linewidth]{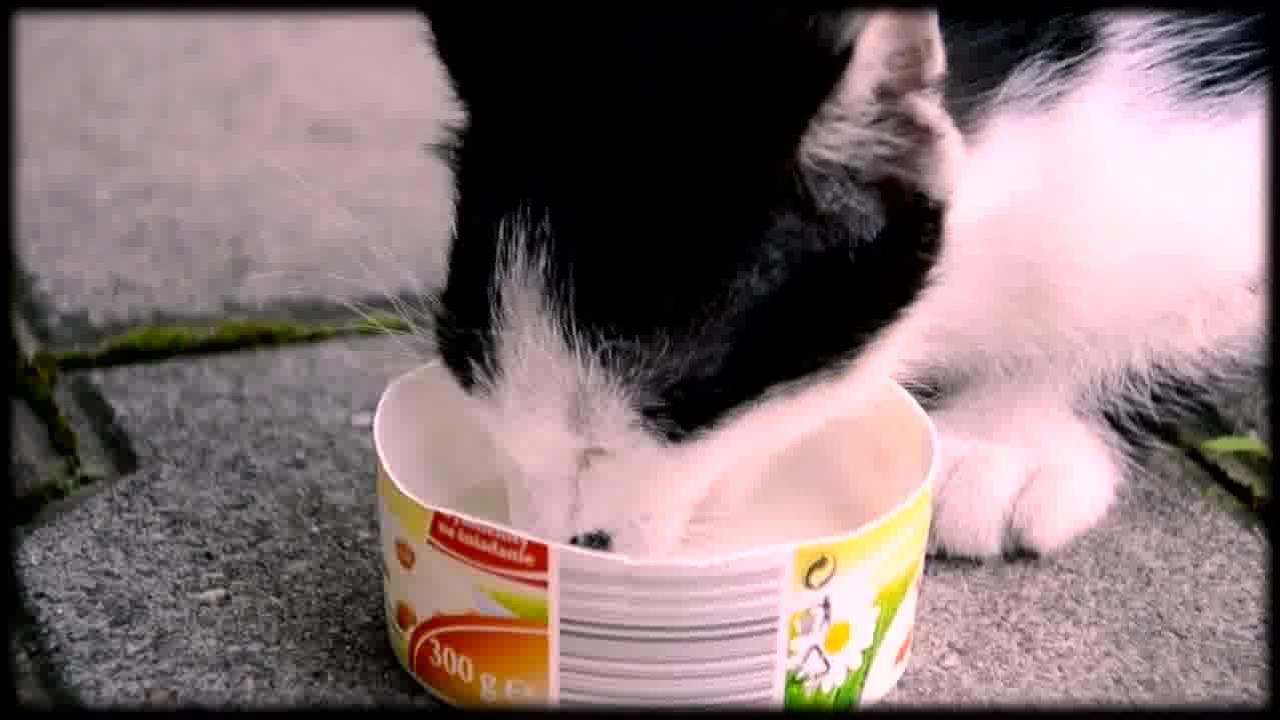}}
\subfigure[]
{\includegraphics[width=0.41\linewidth]{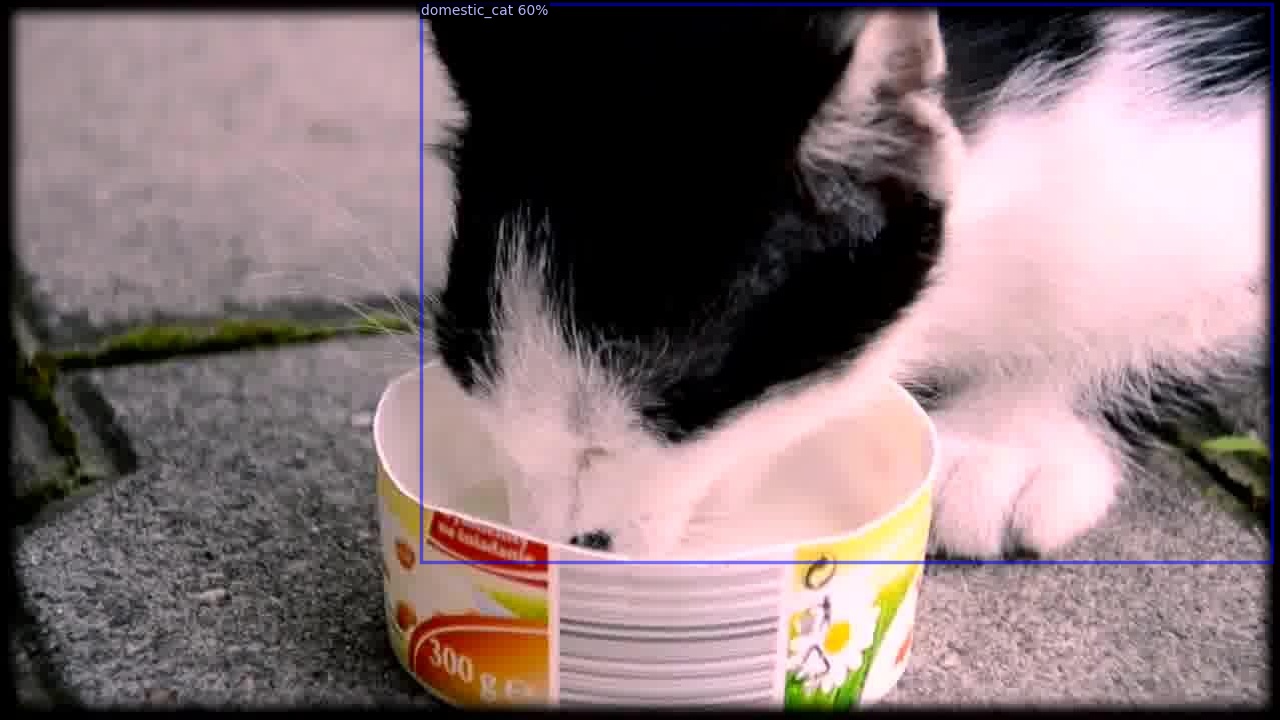}}
% \subfigure[]
% {\includegraphics[width=0.418\linewidth]{Figures/supp/vid_qual/base/000029.JPEG}}
% \subfigure[]
% {\includegraphics[width=0.418\linewidth]{Figures/supp/vid_qual/base/000029.JPEG}}
\hfill \\
\vspace{-20pt}

\subfigure[(a) baseline]
{\includegraphics[width=0.41\linewidth]{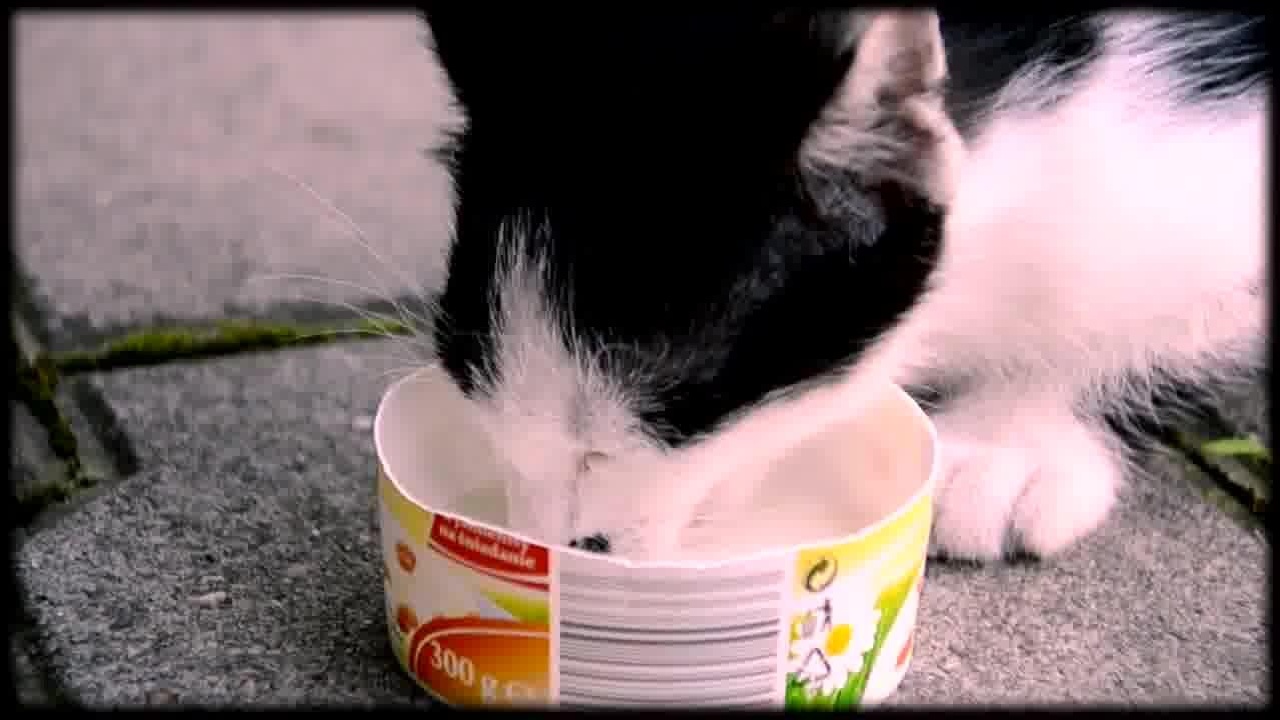}}
\subfigure[(a) CETR]
{\includegraphics[width=0.41\linewidth]{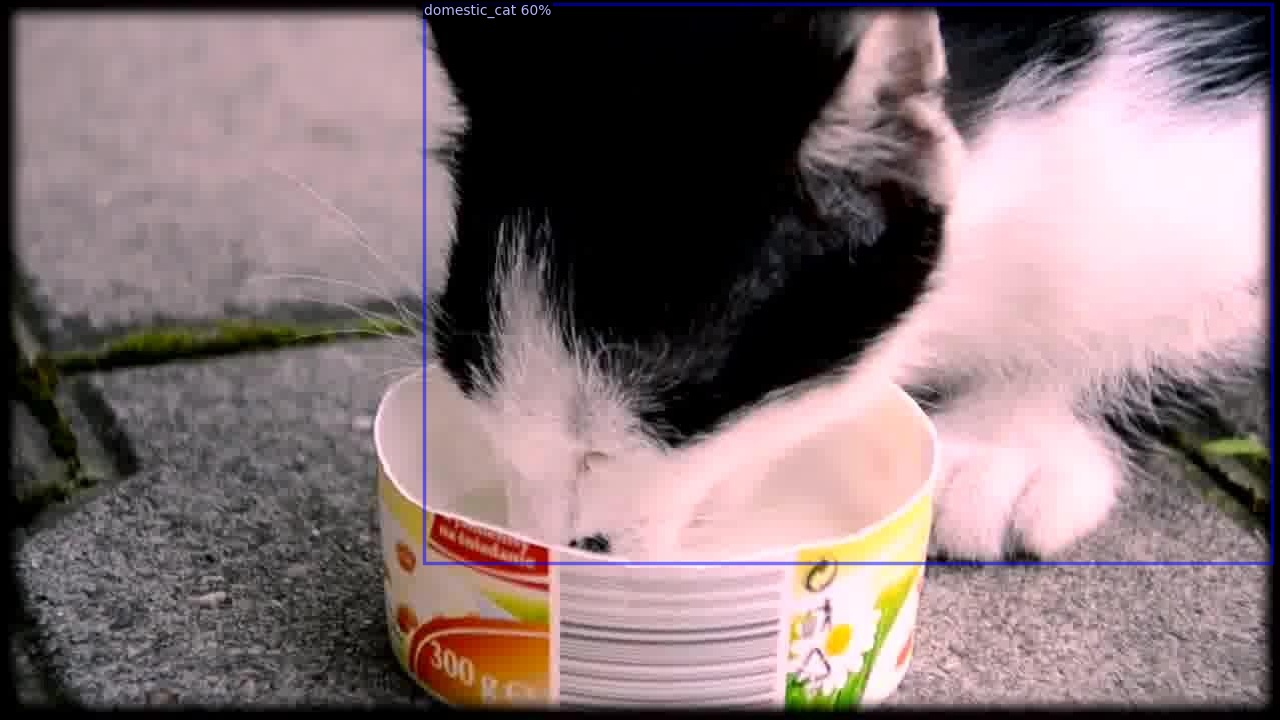}}
\hfill \\

\end{center}
\vspace{-16pt}
\caption{\textbf{Qualitative Results on ImageNet VID dataset~\cite{russakovsky2015imagenet}} comparison between the baseline~\cite{liu2022dab} (left) and our proposed method (right). }%
\label{fig:vid1}\vspace{-10pt}
\end{figure*}

\begin{figure*}[ht!]
\begin{center}
\renewcommand{\thesubfigure}{}

\subfigure[]
{\includegraphics[width=0.41\linewidth]{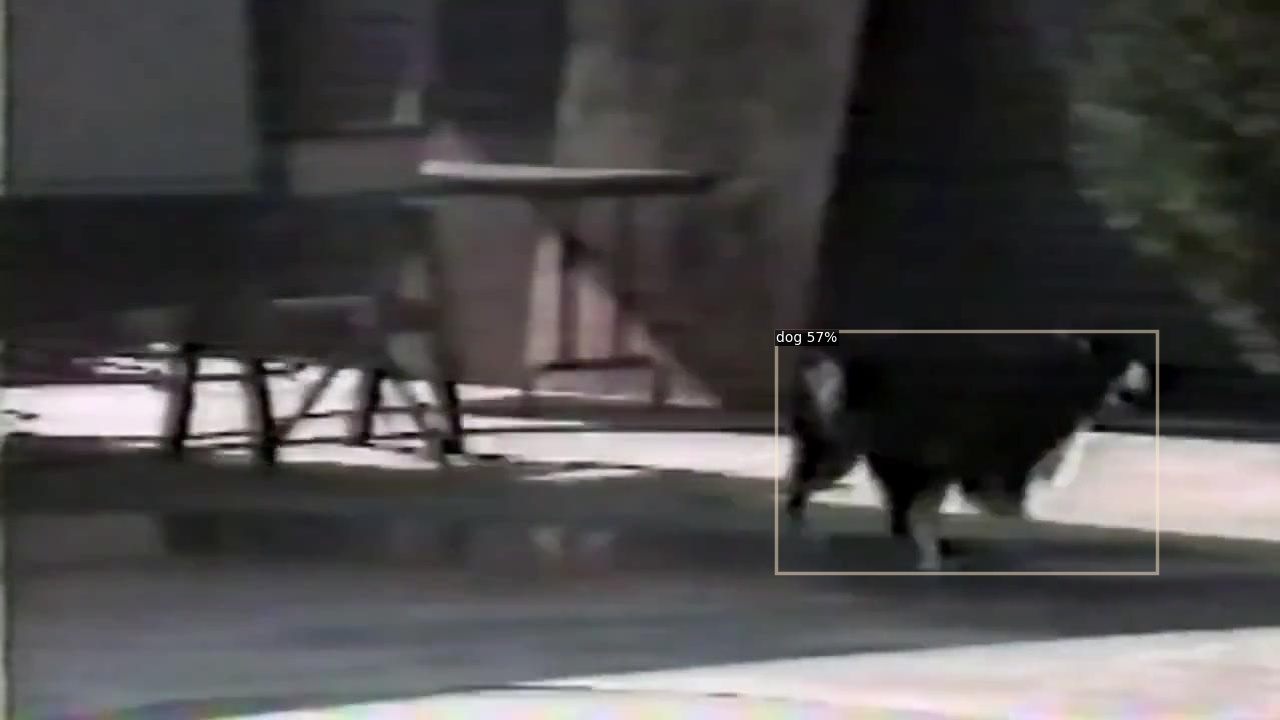}}
\subfigure[]
{\includegraphics[width=0.41\linewidth]{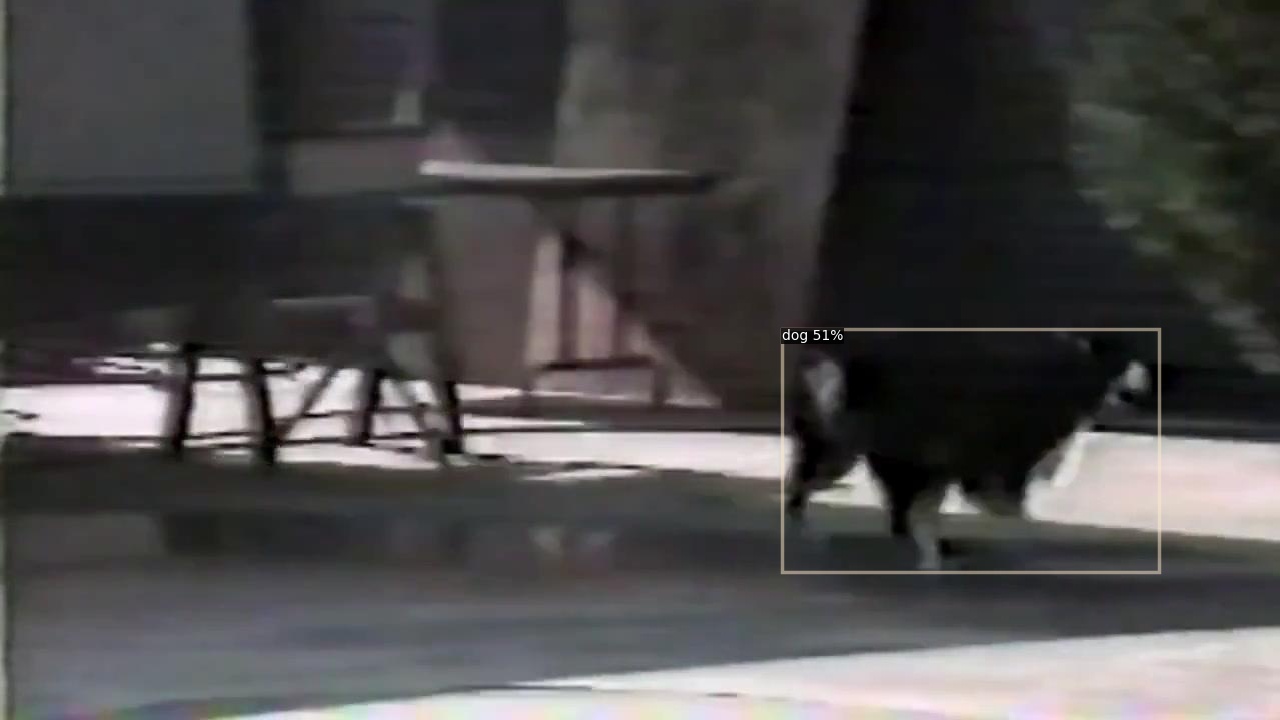}}
% \subfigure[]
% {\includegraphics[width=0.41\linewidth]{Figures/supp/vid_qual/base/000029.JPEG}}
% \subfigure[]
% {\includegraphics[width=0.41\linewidth]{Figures/supp/vid_qual/base/000029.JPEG}}
\hfill \\
\vspace{-20pt}

\subfigure[]
{\includegraphics[width=0.41\linewidth]{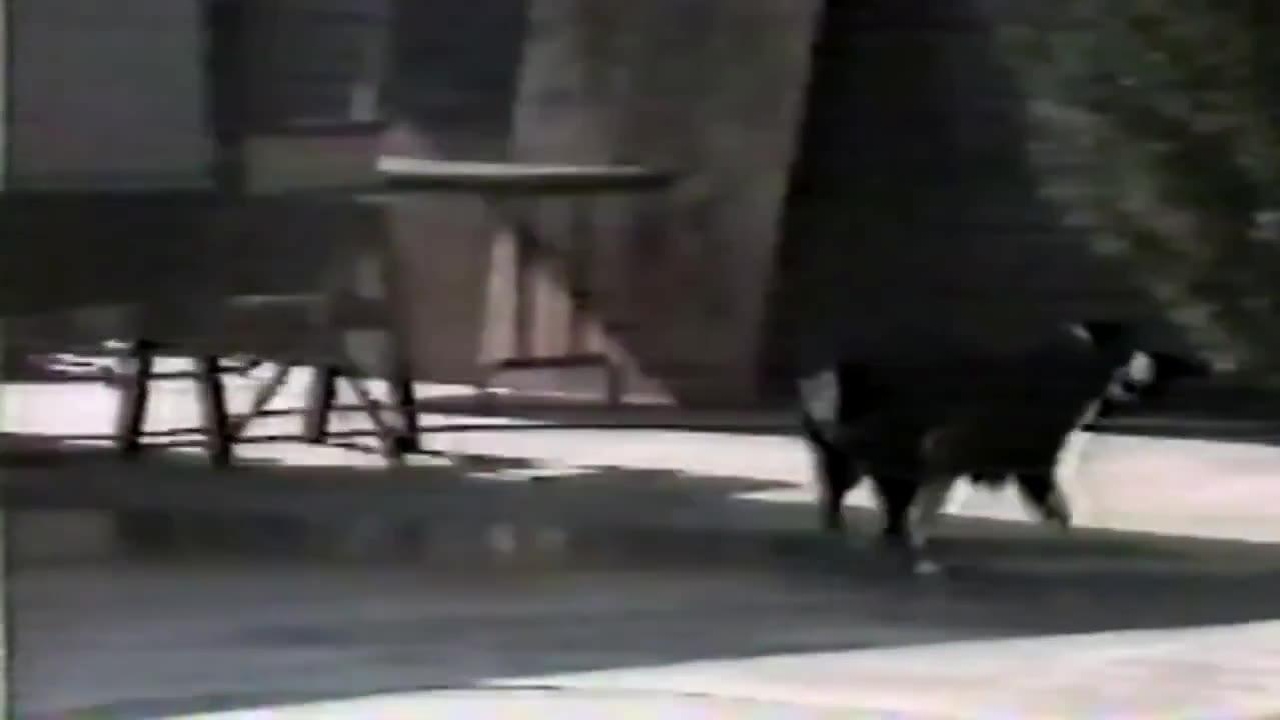}}
\subfigure[]
{\includegraphics[width=0.41\linewidth]{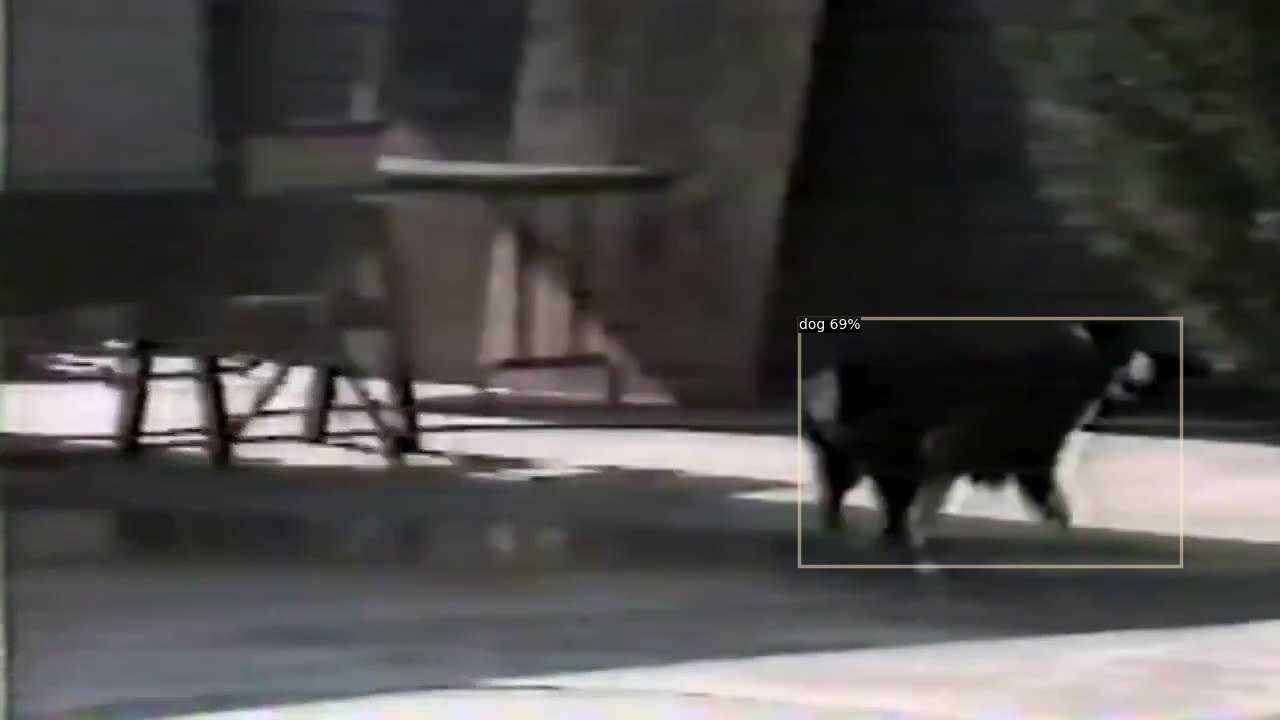}}
% \subfigure[]
% {\includegraphics[width=0.41\linewidth]{Figures/supp/vid_qual/base/000029.JPEG}}
% \subfigure[]
% {\includegraphics[width=0.41\linewidth]{Figures/supp/vid_qual/base/000029.JPEG}}
\hfill \\
\vspace{-20pt}

\subfigure[]
{\includegraphics[width=0.41\linewidth]{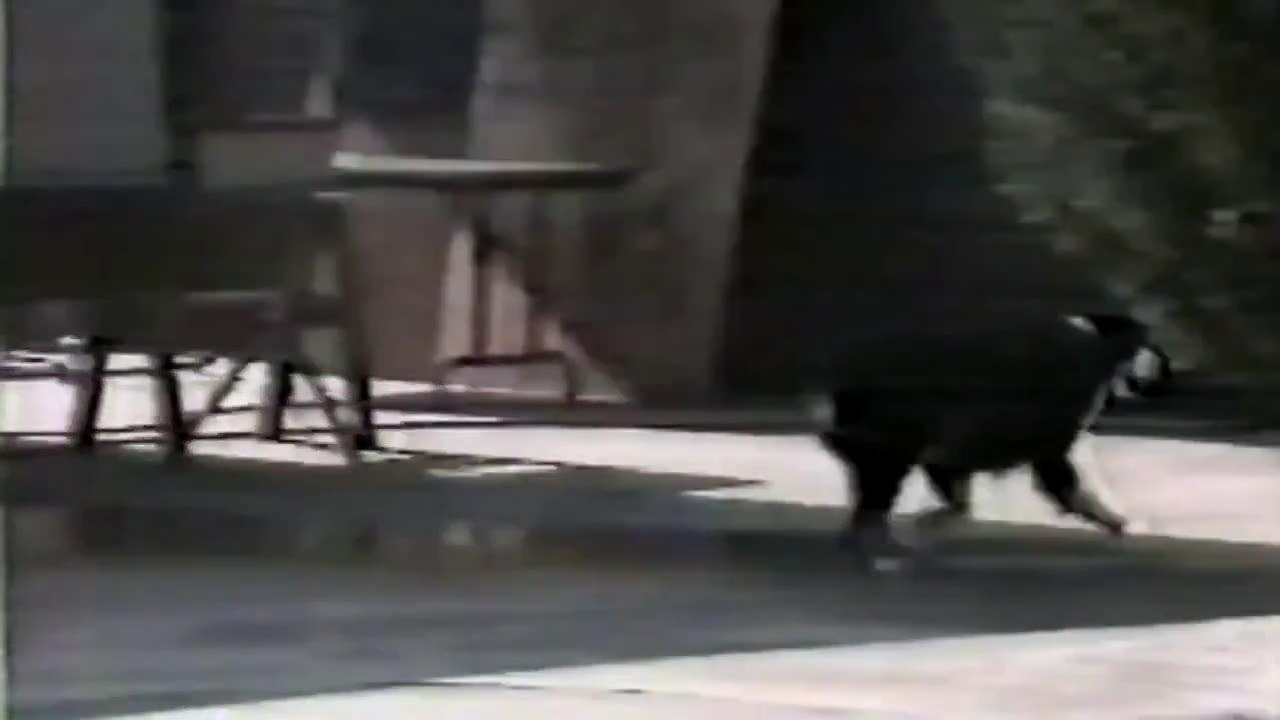}}
\subfigure[]
{\includegraphics[width=0.41\linewidth]{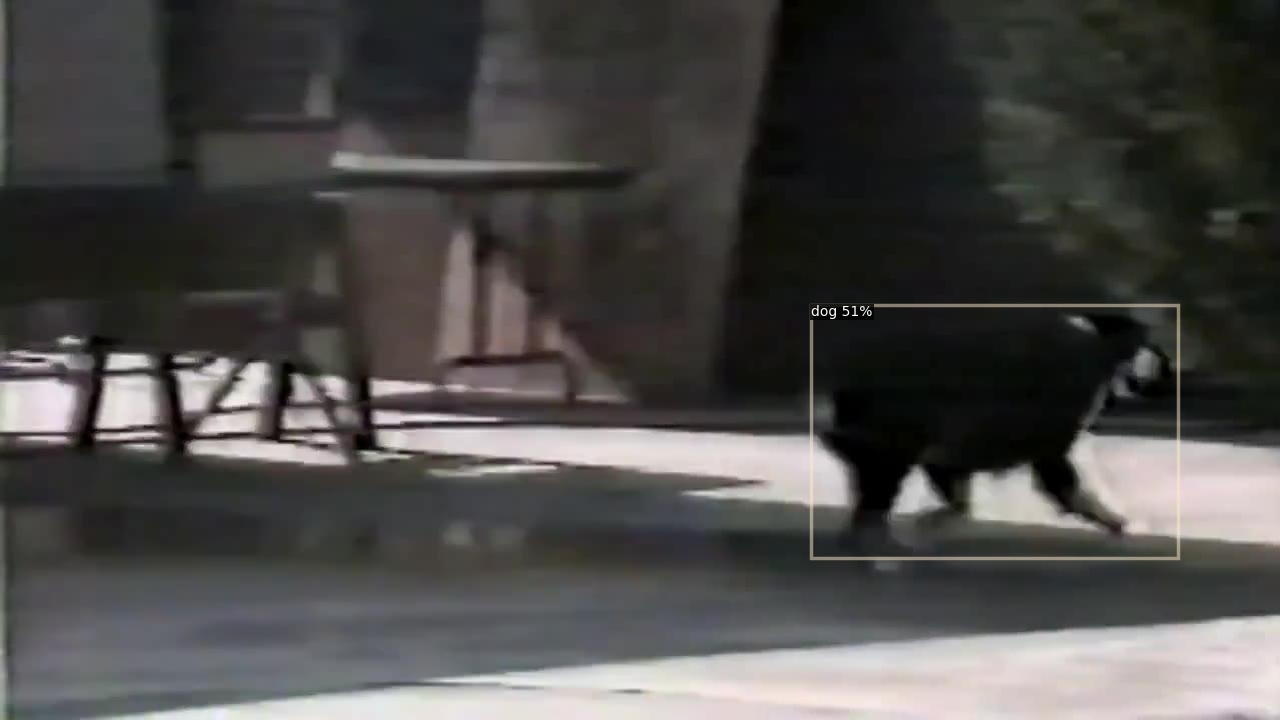}}
% \subfigure[]
% {\includegraphics[width=0.41\linewidth]{Figures/supp/vid_qual/base/000029.JPEG}}
% \subfigure[]
% {\includegraphics[width=0.41\linewidth]{Figures/supp/vid_qual/base/000029.JPEG}}
\hfill \\
\vspace{-20pt}

\subfigure[]
{\includegraphics[width=0.41\linewidth]{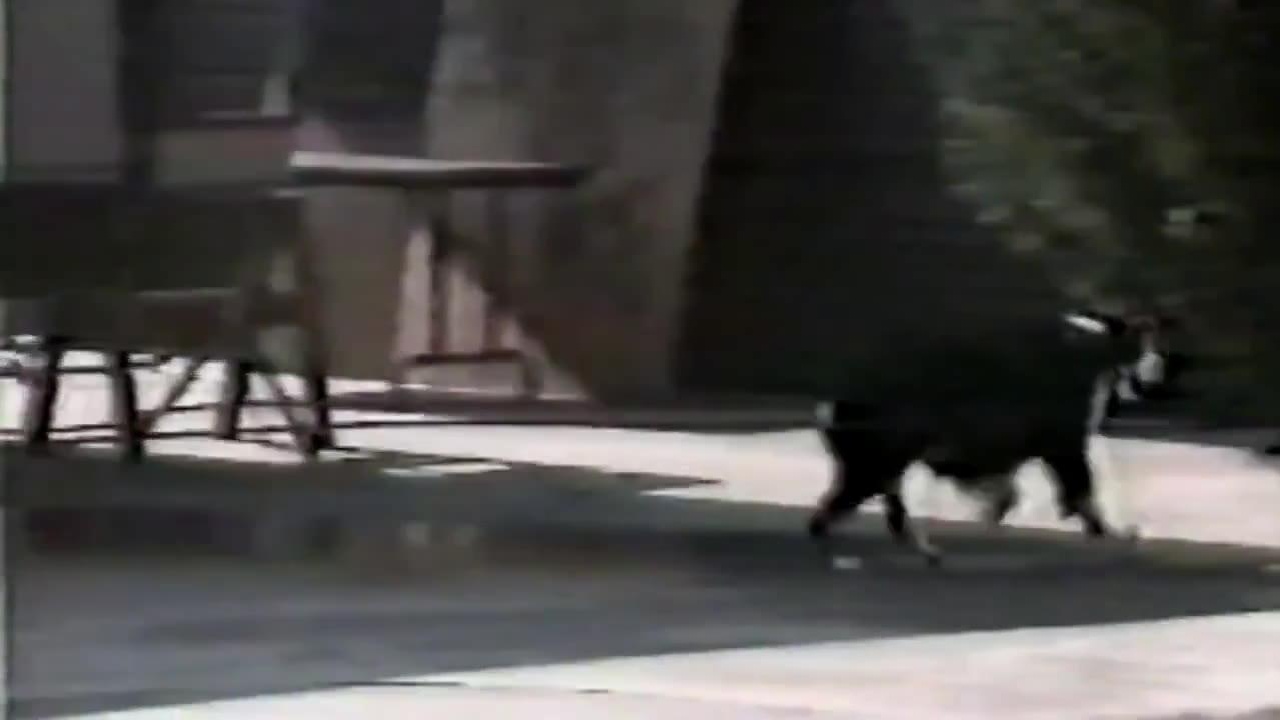}}
\subfigure[]
{\includegraphics[width=0.41\linewidth]{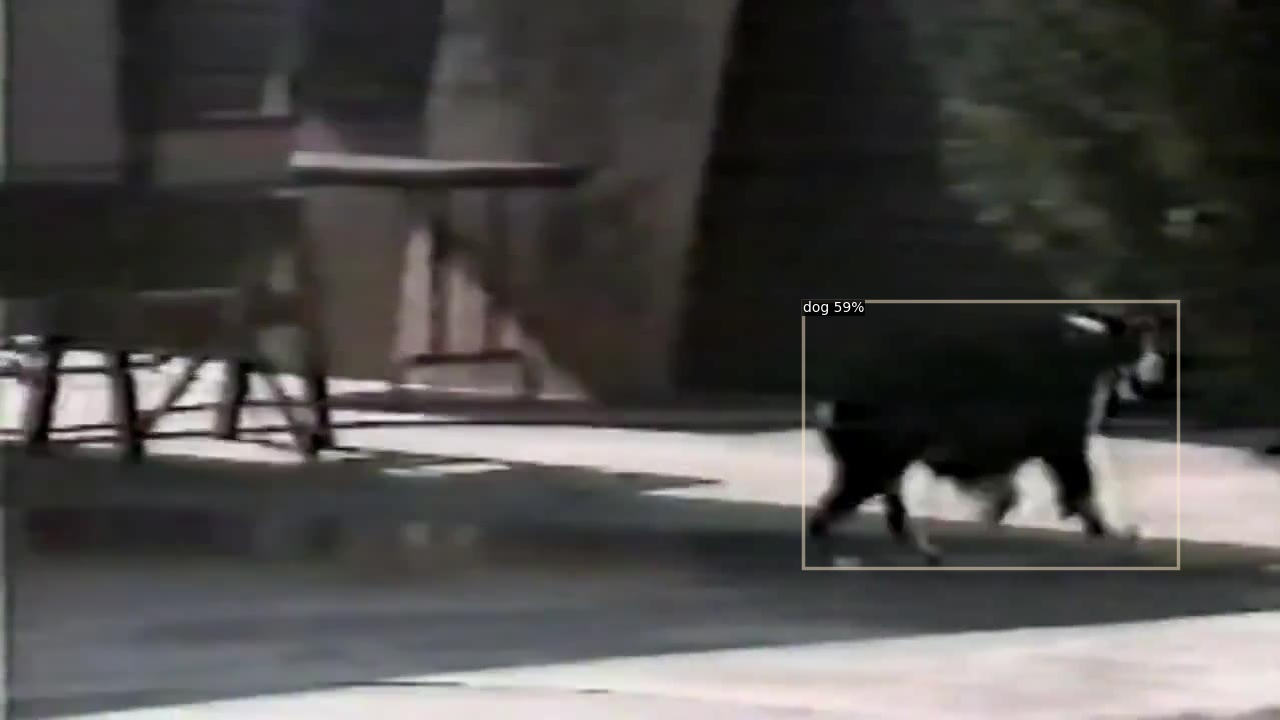}}
% \subfigure[]
% {\includegraphics[width=0.41\linewidth]{Figures/supp/vid_qual/base/000029.JPEG}}
% \subfigure[]
% {\includegraphics[width=0.41\linewidth]{Figures/supp/vid_qual/base/000029.JPEG}}
\hfill \\
\vspace{-20pt}

\subfigure[(a) baseline]
{\includegraphics[width=0.41\linewidth]{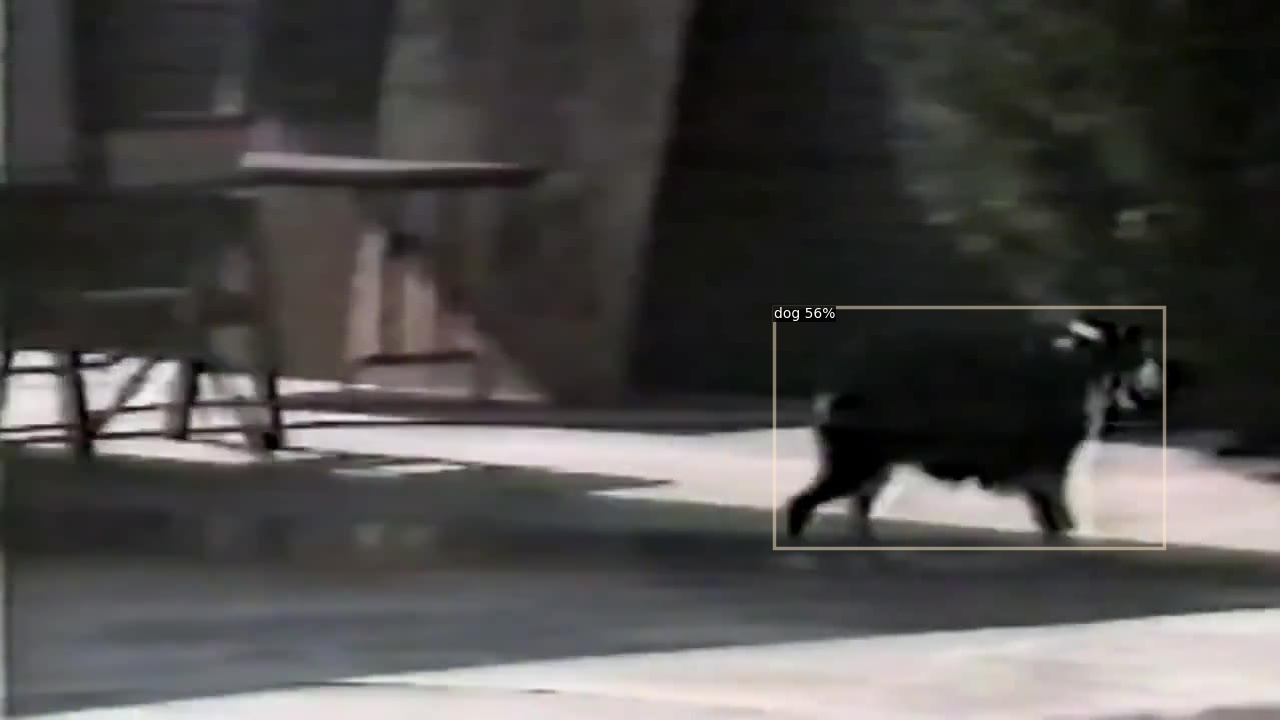}}
\subfigure[(b) CETR]
{\includegraphics[width=0.41\linewidth]{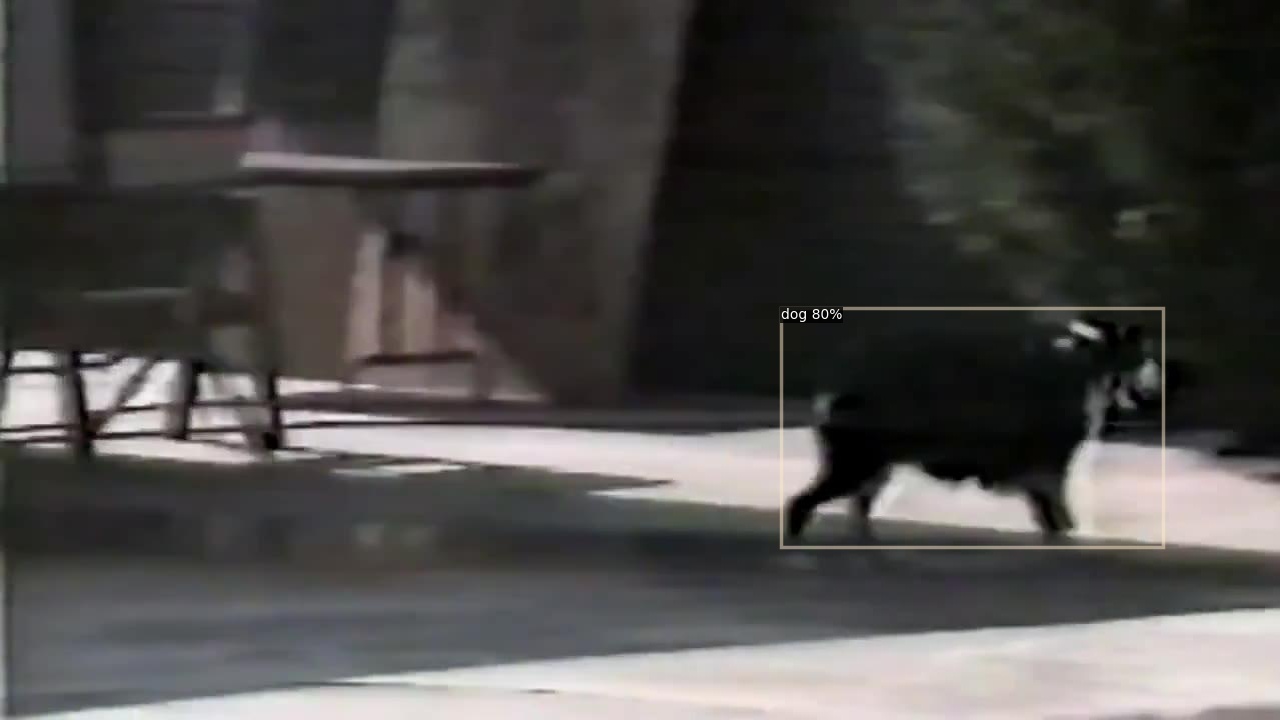}}
\subfigure[]
% {\includegraphics[width=0.41\linewidth]{Figures/supp/vid_qual/base/000029.JPEG}}
% \subfigure[]
% {\includegraphics[width=0.41\linewidth]{Figures/supp/vid_qual/base/000029.JPEG}}
\hfill \\

\end{center}
\vspace{-16pt}
\caption{\textbf{Qualitative Results on ImageNet VID dataset~\cite{russakovsky2015imagenet}} comparison between the baseline~\cite{liu2022dab} (left) and our proposed method (right).  }%
\label{fig:vid2}\vspace{-10pt}
\end{figure*}

\begin{figure*}[ht!]
\begin{center}
\renewcommand{\thesubfigure}{}

\subfigure[]
{\includegraphics[width=0.41\linewidth]{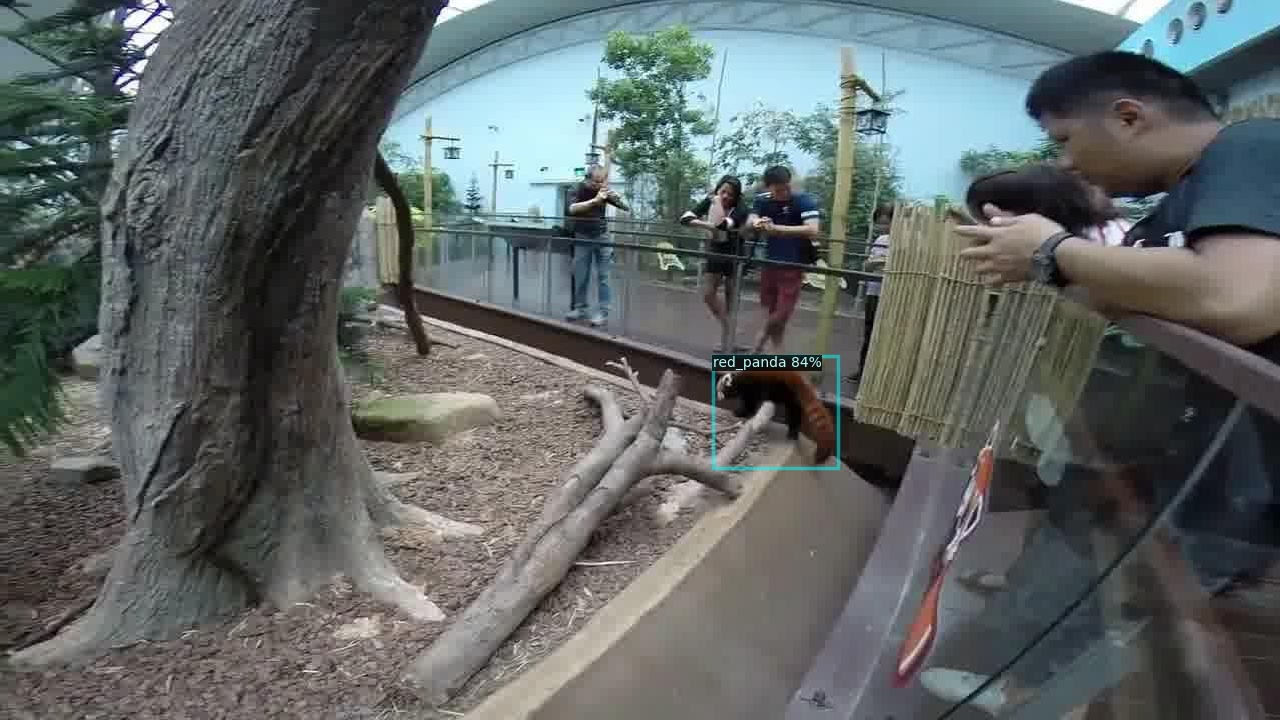}}
\subfigure[]
{\includegraphics[width=0.41\linewidth]{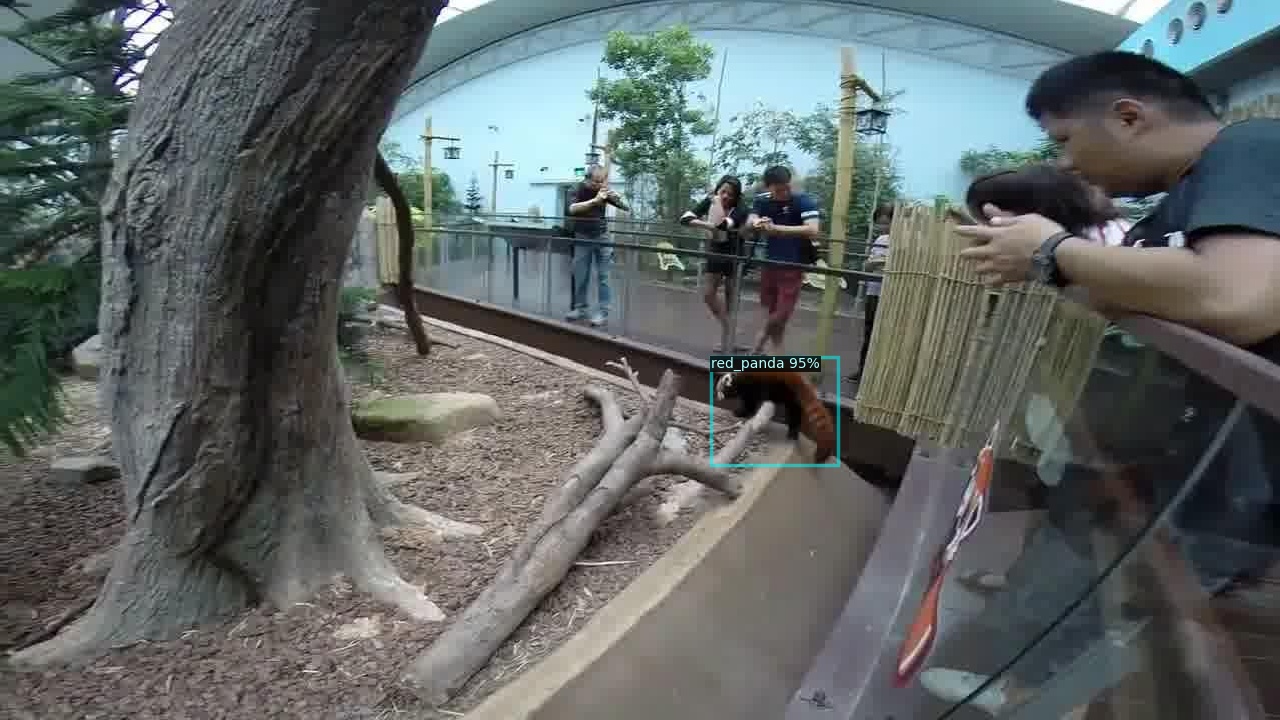}}
\hfill \\
\vspace{-20pt}

\subfigure[]
{\includegraphics[width=0.41\linewidth]{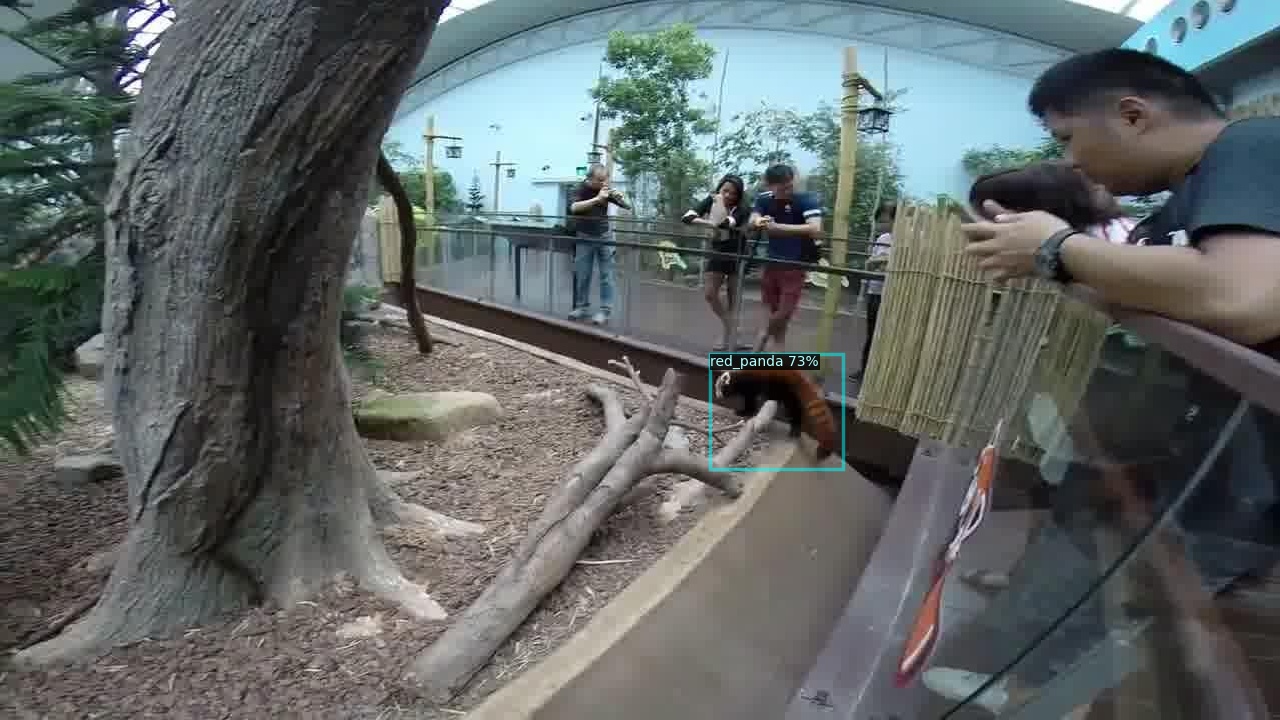}}
\subfigure[]
{\includegraphics[width=0.41\linewidth]{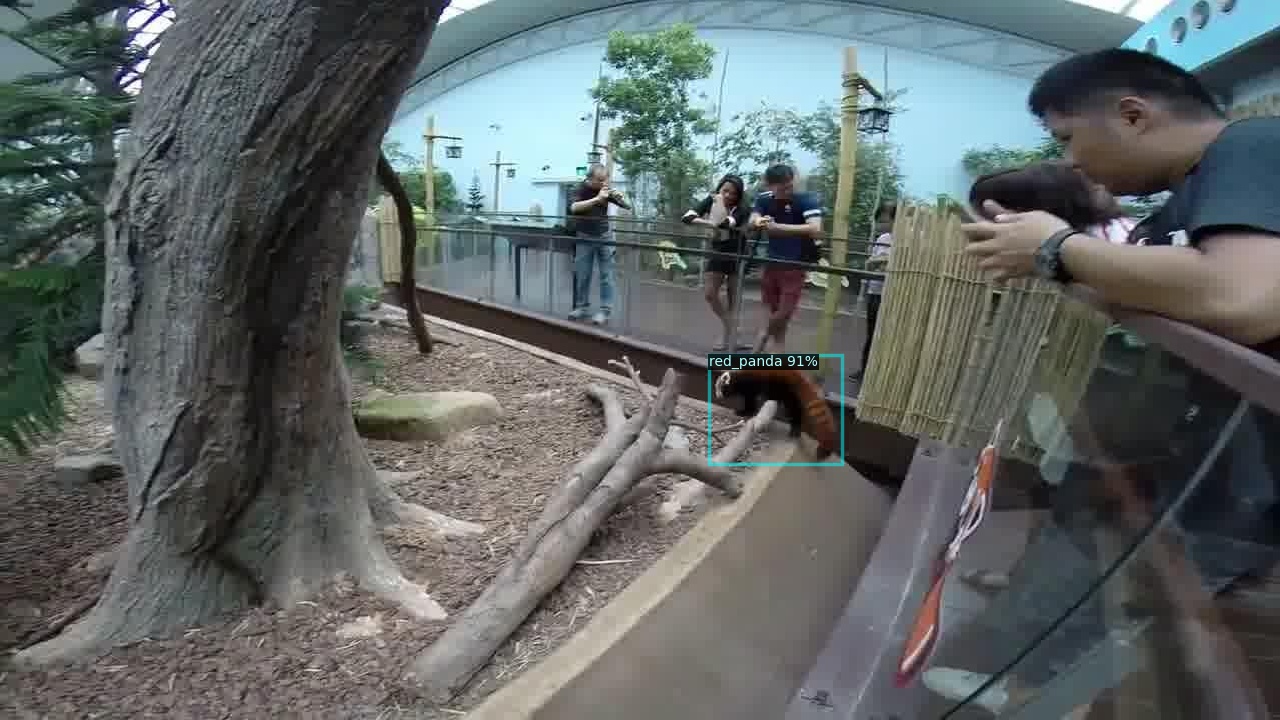}}

\hfill \\
\vspace{-20pt}

\subfigure[]
{\includegraphics[width=0.41\linewidth]{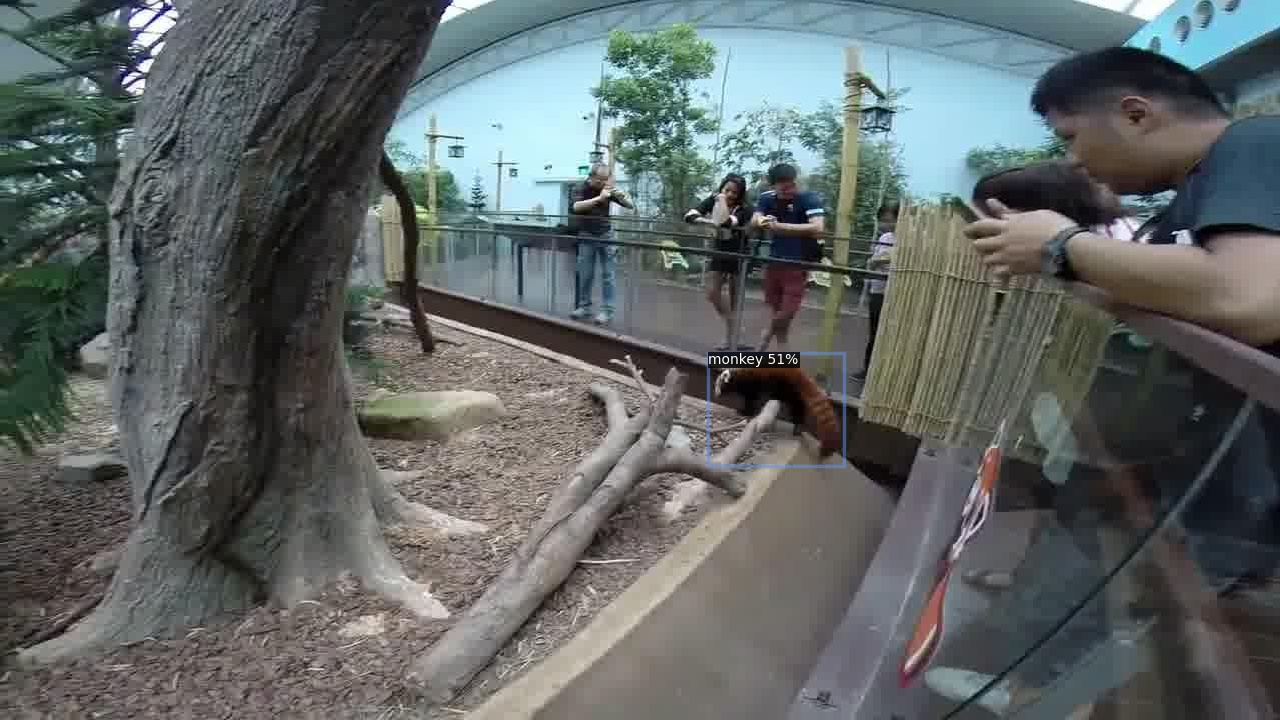}}
\subfigure[]
{\includegraphics[width=0.41\linewidth]{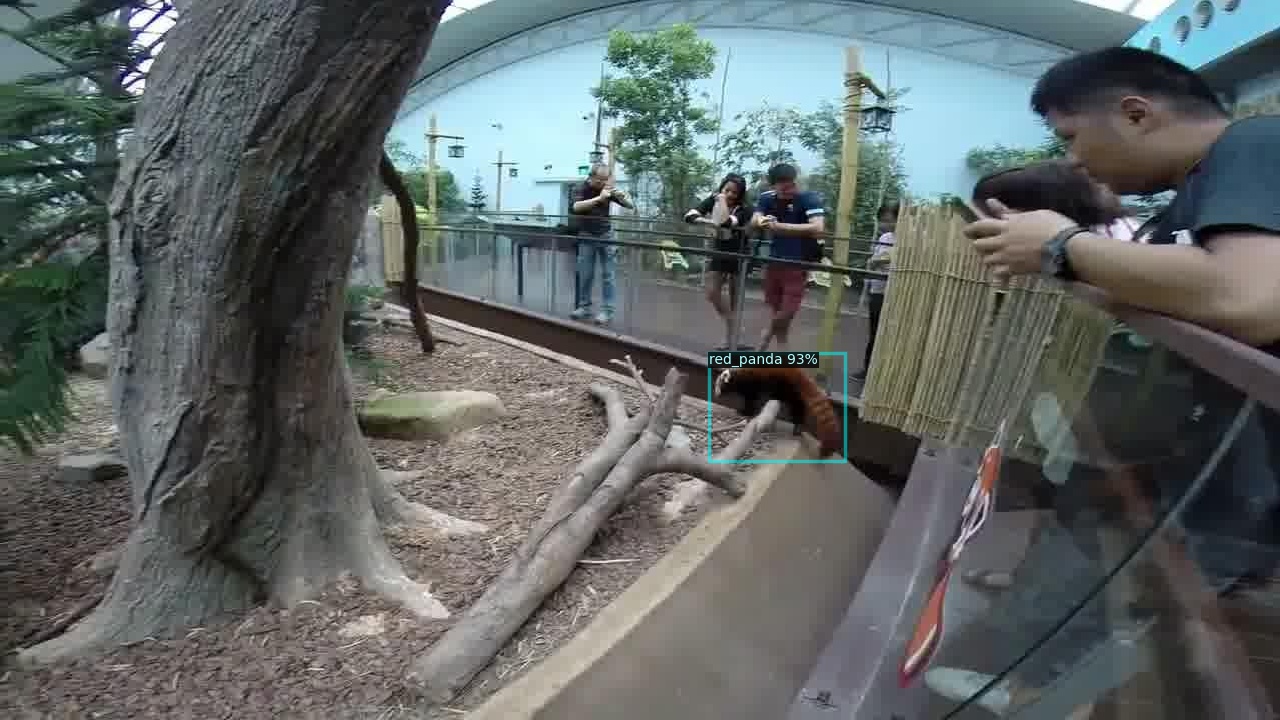}}
\hfill \\
\vspace{-20pt}

\subfigure[]
{\includegraphics[width=0.41\linewidth]{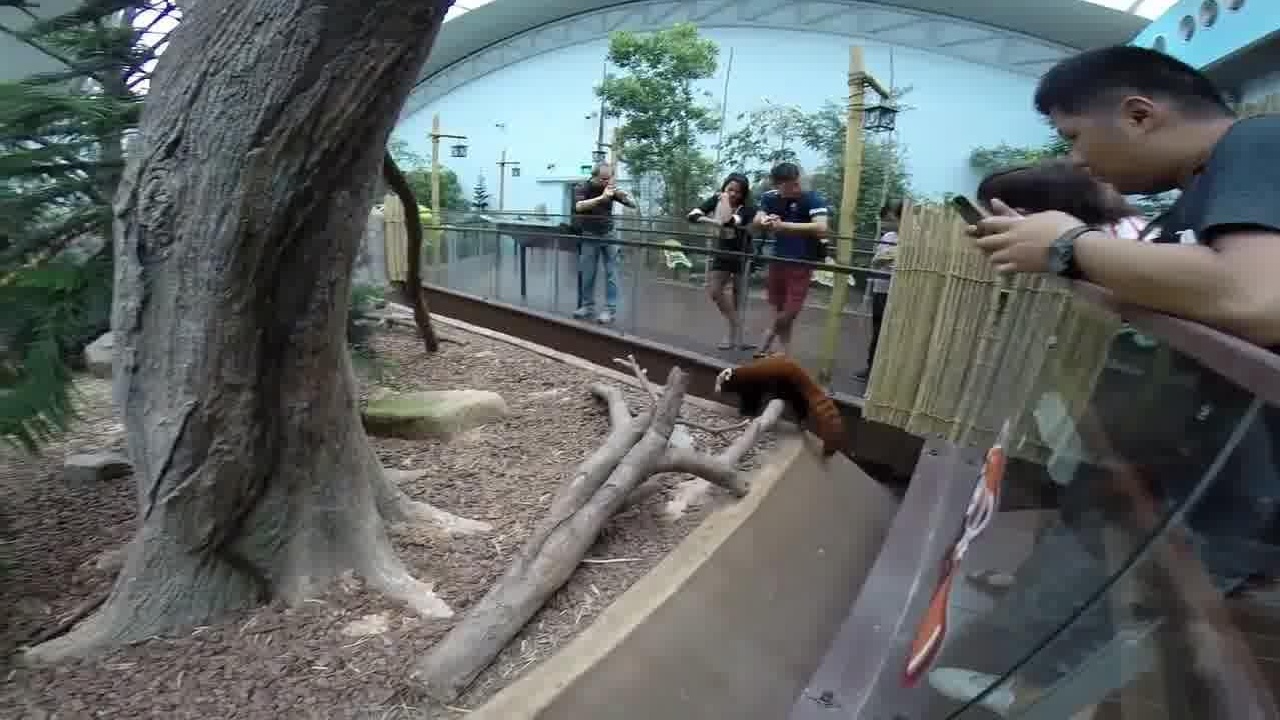}}
\subfigure[]
{\includegraphics[width=0.41\linewidth]{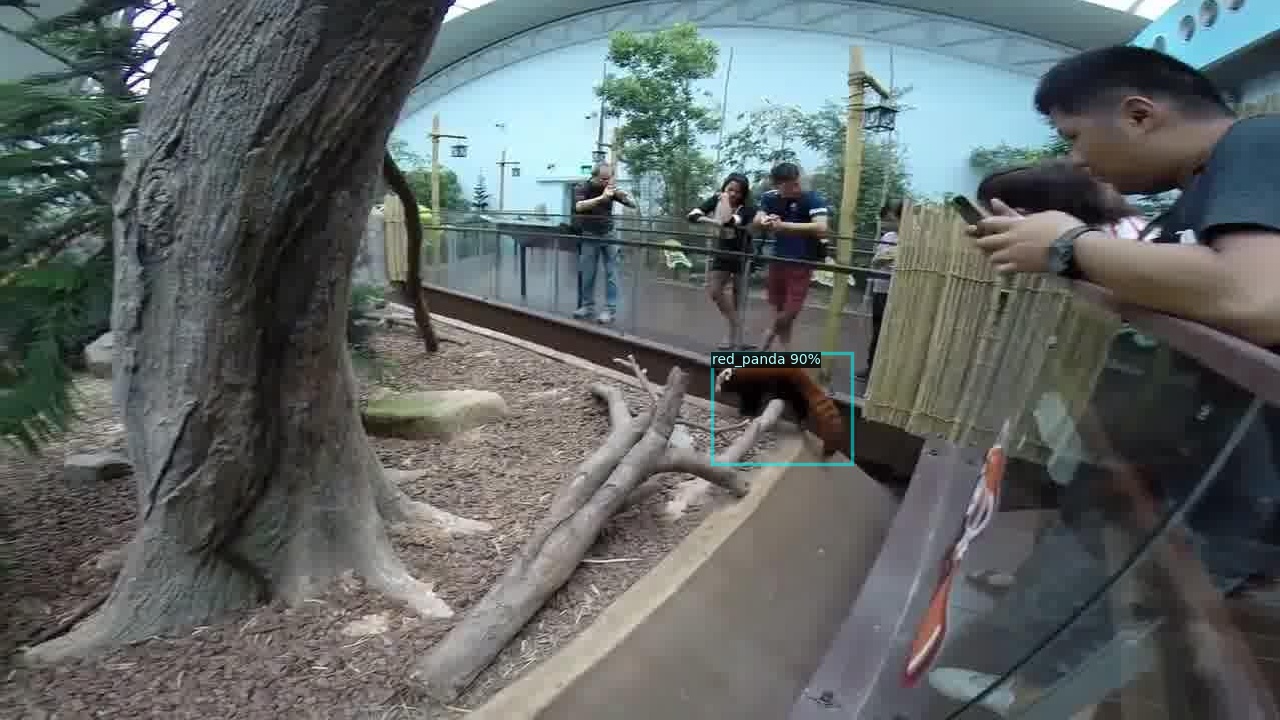}}

\hfill \\
\vspace{-20pt}

\subfigure[(a) baseline]
{\includegraphics[width=0.41\linewidth]{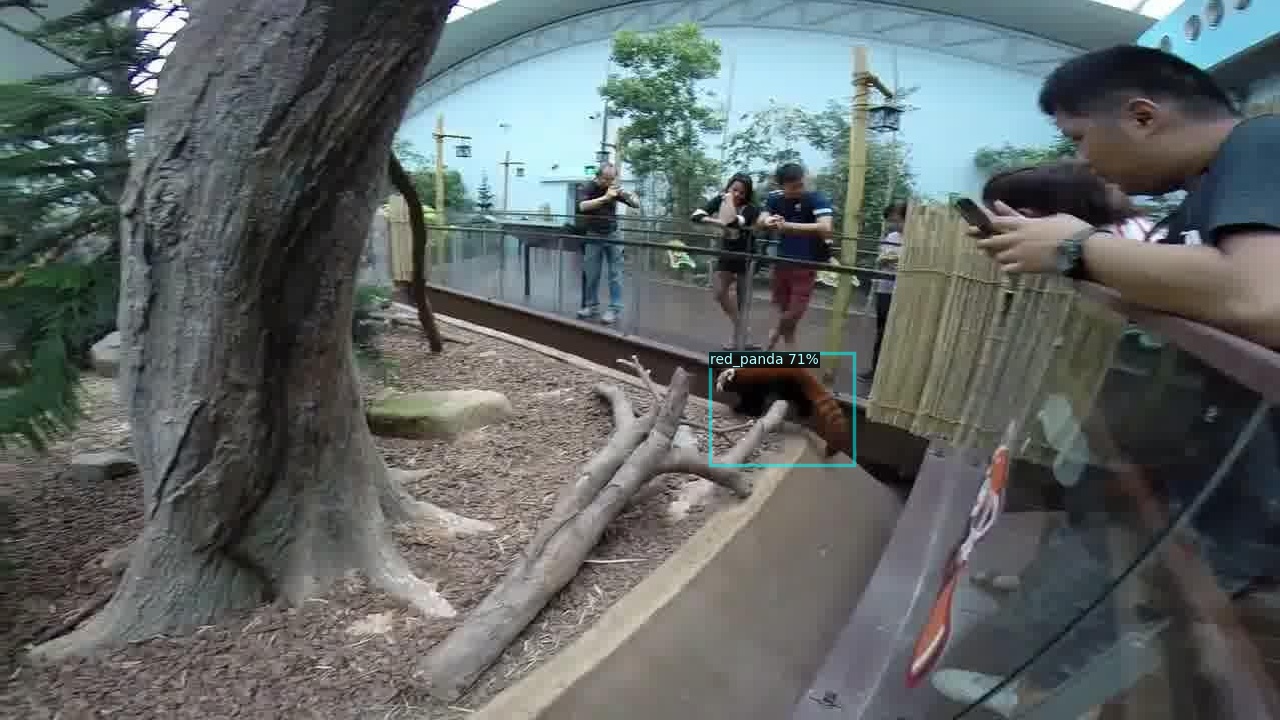}}
\subfigure[(b) CETR]
{\includegraphics[width=0.41\linewidth]{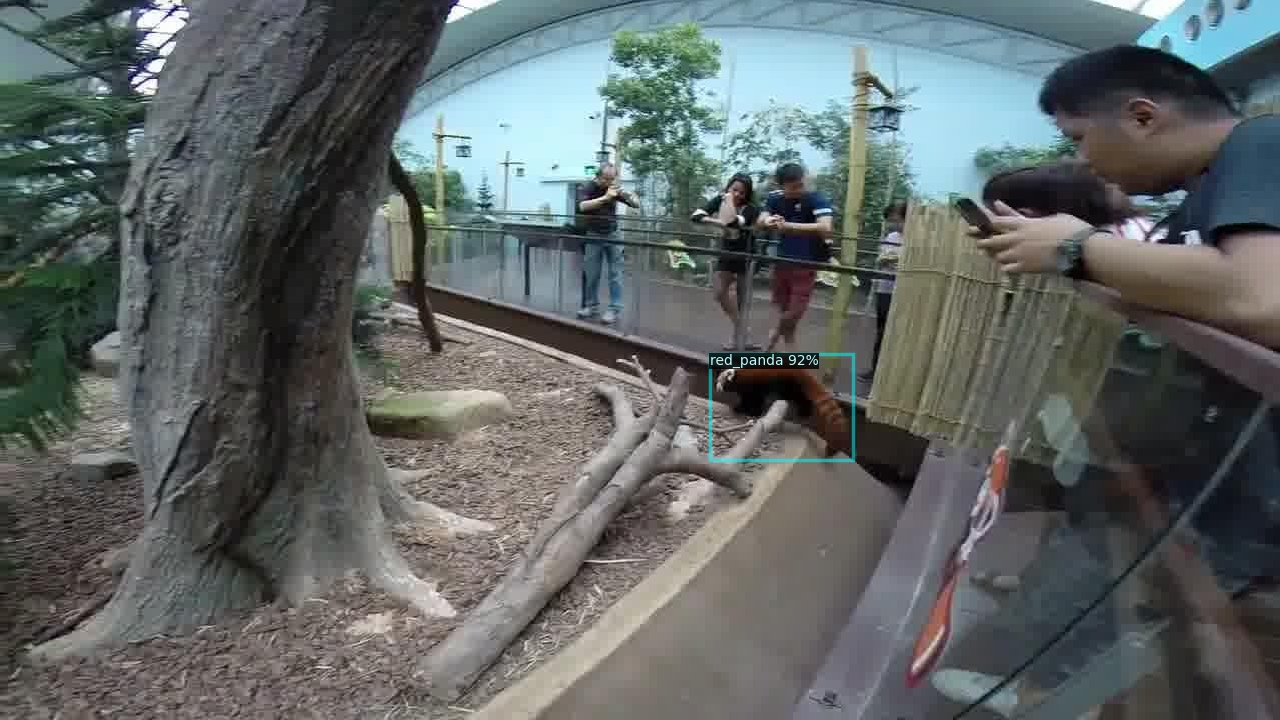}}

\hfill \\

\end{center}
\vspace{-16pt}
\caption{\textbf{Qualitative Results on ImageNet VID dataset~\cite{russakovsky2015imagenet}} comparison between the baseline~\cite{liu2022dab} (left) and our proposed method (right). }%
\label{fig:vid3}\vspace{-10pt}
\end{figure*}

\begin{figure*}[ht!]
\begin{center}
\renewcommand{\thesubfigure}{}

\subfigure[]
{\includegraphics[width=0.41\linewidth]{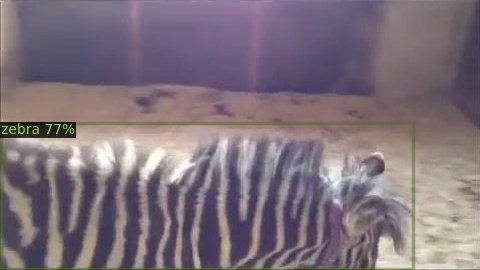}}
\subfigure[]
{\includegraphics[width=0.41\linewidth]{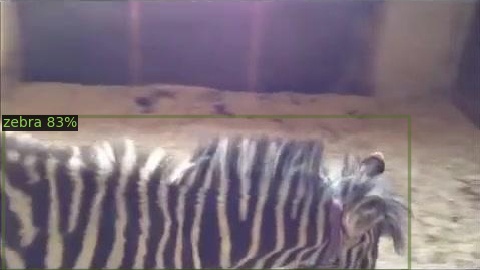}}
\hfill \\
\vspace{-20pt}

\subfigure[]
{\includegraphics[width=0.41\linewidth]{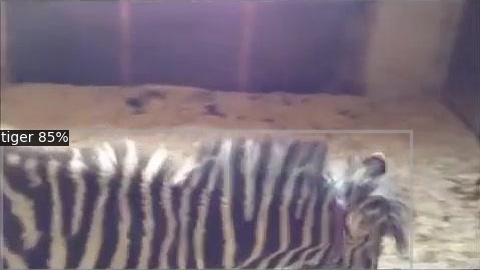}}
\subfigure[]
{\includegraphics[width=0.41\linewidth]{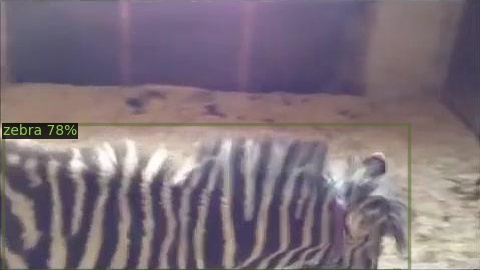}}

\hfill \\
\vspace{-20pt}

\subfigure[]
{\includegraphics[width=0.41\linewidth]{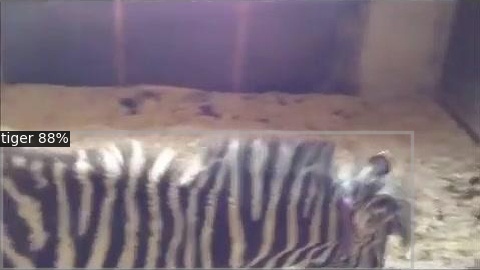}}
\subfigure[]
{\includegraphics[width=0.41\linewidth]{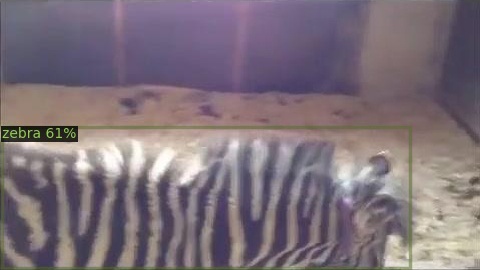}}
\hfill \\
\vspace{-20pt}

\subfigure[]
{\includegraphics[width=0.41\linewidth]{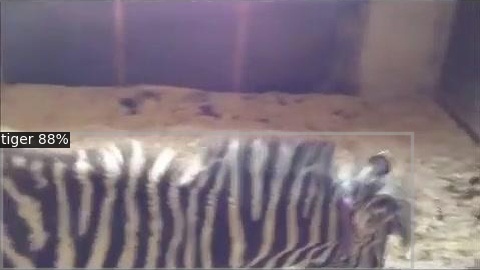}}
\subfigure[]
{\includegraphics[width=0.41\linewidth]{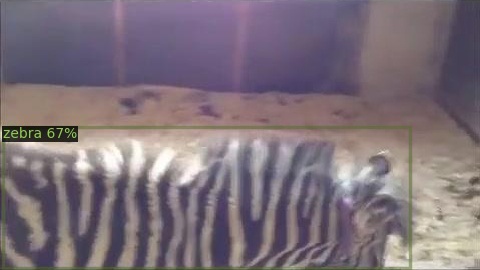}}

\hfill \\
\vspace{-20pt}

\subfigure[(a) baseline]
{\includegraphics[width=0.41\linewidth]{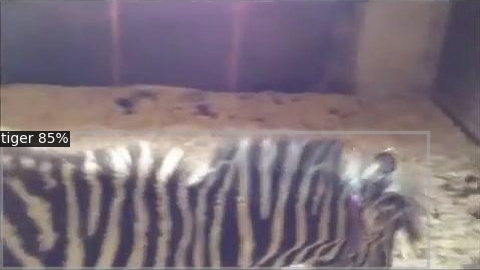}}
\subfigure[(b) CETR]
{\includegraphics[width=0.41\linewidth]{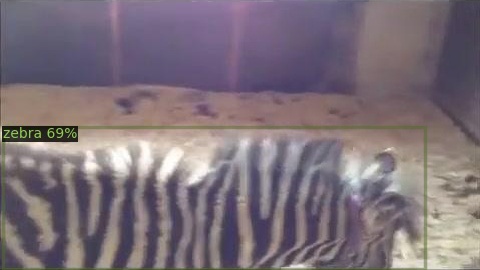}}

\hfill \\

\end{center}
\vspace{-16pt}
\caption{\textbf{Qualitative Results on ImageNet VID dataset~\cite{russakovsky2015imagenet}} comparison between the baseline~\cite{liu2022dab} (left) and our proposed method (right).}%
\label{fig:vid4}\vspace{-10pt}
\end{figure*}

\setlabel{C}{sec:qual}
In this section, we provide additional qualitative results of the baseline, i.e., DAB-DETR~\cite{liu2022dab}, and our proposed framework on the CityCam dataset~\cite{zhang2017understanding} in Figure~\ref{fig:citycam} and ImageNet VID dataset~\cite{russakovsky2015imagenet} from Figure~\ref{fig:vid1} to Figure~\ref{fig:vid4}. Our method provides better detection results than the baseline. From the depicted figures, we notice that the baseline, single frame detection, misses objects between frames, while CETR confidently predicts objects due to the ability of context memory module.

\newpage

\newpage
\begin{algorithm}[h]
\caption{Training-time Memory Update}\label{alg:Training}
\LinesNumbered
\KwIn{\\
\quad image features of time step $t$: \\
\quad \quad $\mathcal{F}^{(t)}\in \mathbb{R}^{H\cdot W\times d}$, \\
\quad ground-truth instance boxes of time step $t$: \\
\quad \quad $B^{(t)}=\{ x_n, y_n, h_n, w_n, c_n \}^{N}_{n=1}$, \\
\quad class-wise memory of time step $t-1$: \\
\quad \quad $M^{(t-1)} = \{ m_{c,k} \}^{C,K}_{c,k=1} \in \mathbb{R}^{C\cdot K\times d}$}
\KwOut{\\
\quad updated class-wise memory $M^{(t)}$}
{\tcc{ Superscript symbols $n$, $c$, and \\
$k$ mean index of instance, class, and prototype.}}
$\tilde{\mathcal{F}}^{(t)} \gets \text{RoIAlign}(\mathcal{F}^{(t)}, B^{(t)}) = \{ f^{c_n}_n \}^N_{n=1} $

 %Setting temperature $\tau$ as a constant.

\For{$n=1\textrm{ to } N$}{
     $k_n \gets \text{argmax}_k\{ {f^{c_n}_n}^\top m_{c_n,k} \}^K_{k=1}$ \\
     $m_{c_n,k_n} \gets \alpha m_{c_n,k_n} + (1 - \alpha) f^{c_n}_n$ \\
     {\small \tcp{$\alpha$ is a momentum coefficient.}}
}
${{M}}^{(t)}\gets {{M}}^{(t-1)}$  \\
$n=1,2,...,N; \ c=1,2,...,C; \ k=1,2..,K$
\end{algorithm}
\begin{algorithm}[h!]
\caption{Test-time Memory Adaptation}\label{alg:TTA}
\LinesNumbered
\KwIn{\\
\quad image features of time step $t$: \\
\quad \quad $\mathcal{F}^{(t)}\in \mathbb{R}^{H\cdot W\times d}$, \\
\quad predicted instance boxes of time step $t$: \\
\quad \quad $B^{(t)}=\{ x_n, y_n, h_n, w_n, c_n \}^{N}_{n=1}$, \\
\quad test-time class-wise memory of time step $t-1$: \\
\quad \quad $M'^{(t-1)} = \{ m'_{c,k} \}^{C,K}_{c,k=1} \in \mathbb{R}^{C\cdot K\times d}$ \\
\quad test-time class-wise scale of time step $t-1$: \\
\quad \quad $\mathcal{I}^{(t-1)} = \{  \frac{1}{i_{c,k}} \}^{C,K}_{c,k=1} \in \mathbb{R}^{C\cdot K\times 1}$ \\
\quad training-time class-wise memory: \\
\quad \quad $M = \{ m_{c,k} \}^{C,K}_{c,k=1} \in \mathbb{R}^{C\cdot K\times d}$}
\KwOut{\\
\quad test-time class-wise scale $\mathcal{I}^{(t)}$  \\
\quad adapted class-wise memory $M''$
}

% {\tcc{ Superscript symbols $n$, $c$, and \\
% $k$ mean index of instance, class, and prototype.}}
$\tilde{\mathcal{F}}^{(t)} \gets \text{RoIAlign}(\mathcal{F}^{(t)}, B^{(t)}) = \{ f^{c_n}_n \}^N_{n=1} $

 %Setting temperature $\tau$ as a constant.

\For{$n=1\textrm{ to } N$}{
     $k_n \gets \text{argmax}_k\{ {f^{c_n}_n}^\top m_{c_n,k} \}^K_{k=1}$ \\

     % $m'_{c,k} \gets \frac{1}{i_{c,k}}\,( m'_{c,k} \, + \, \sum^{J}_{j=1}f_{j}^{c})$ \\
     $m'_{c_n,k_n} \gets m'_{c_n,k_n} + f_{n}^{c_n}$ \\
         
     $i_{c_n,k_n} \gets i_{c_n,k_n} + 1$ \\
}
$M'^{(t)}\gets M'^{(t-1)}$  \\
$\mathcal{I}^{(t)}\gets \mathcal{I}^{(t-1)}$  \\
$M''\gets \mathcal{I}^{(t)} \odot M'^{(t)}$  \\
$M''\gets \beta M + (1-\beta) M''$  \\
{\small \tcp{$\beta$ is the weight of the source domain.}}
$n=1,2,...,N; \ c=1,2,...,C; \ k=1,2..,K$
\end{algorithm}

\newpage

\end{document}